\documentclass{article}

\usepackage{microtype}
\usepackage{graphicx}
\usepackage{subfigure}
\usepackage{booktabs}
 \usepackage{bbm} 

\usepackage{hyperref}

\usepackage[accepted]{icml2024}

\usepackage{amsmath}
\usepackage{amssymb}
\usepackage{mathtools}
\usepackage{amsthm}

\usepackage[capitalize,noabbrev]{cleveref}

\theoremstyle{plain}
\newtheorem{theorem}{Theorem}[section]

\newtheorem{lemma}[theorem]{Lemma}
\newtheorem{corollary}[theorem]{Corollary}
\theoremstyle{definition}
\newtheorem{definition}[theorem]{Definition}

\theoremstyle{remark}
\newtheorem{remark}[theorem]{Remark}

\usepackage[textsize=tiny]{todonotes}

\icmltitlerunning{Trustworthy Actionable Perturbations}

\begin{document}

\twocolumn[
\icmltitle{Trustworthy Actionable Perturbations}



\icmlsetsymbol{equal}{*}

\begin{icmlauthorlist}
\icmlauthor{Jesse Friedbaum}{AM,ECE}
\icmlauthor{Sudarshan Adiga}{ECE}
\icmlauthor{Ravi Tandon}{AM,ECE}

\end{icmlauthorlist}

\icmlaffiliation{AM}{Program in Applied Mathematics, University of Arizona, Tucson, AZ, USA}
\icmlaffiliation{ECE}{Department of Electrical and Computer Engineering, University of Arizona, Tucson, AZ, USA}

\icmlcorrespondingauthor{Jesse Friedbaum}{friedbaum@math.arizona.edu}
\icmlcorrespondingauthor{Ravi Tandon}{tandonr@arizona.edu}

\icmlkeywords{Machine Learning, ICML}

\vskip 0.3in
]



\printAffiliationsAndNotice{} 

\begin{abstract}
\textit{Counterfactuals}, or modified inputs that lead to a different outcome, are an important tool for understanding the logic used by machine learning classifiers and how to change an undesirable classification.  Even if a counterfactual changes a classifier's decision, however, it may not affect the true underlying class probabilities, i.e. the  counterfactual may act like an adversarial attack and ``fool'' the classifier.  We propose a new framework for creating modified inputs that change the true underlying probabilities in a beneficial way which we call \textit{Trustworthy Actionable Perturbations} (TAP).  This includes a novel verification procedure to ensure that TAP change the true class probabilities instead of acting adversarially.  Our framework also includes new cost, reward, and goal definitions that are better suited to effectuating change in the real world.  We present PAC-learnability results for our verification procedure and theoretically analyze our new method for measuring reward.  We also develop a methodology for creating TAP and compare our results to those achieved by previous counterfactual methods.
\end{abstract}

\section{Introduction}
\label{intro}

As machine learning (ML) classifiers have experienced widespread adoption in applications that have an out-sized impact on individuals' lives (such as credit lending \citep{banking_APP}, college admissions \citep{admissions_APP} and  healthcare \citep{icu_APP}), the need to understand classifiers' decision making and how to avoid undesirable classifications has become increasingly important.  One of the most important tools for filling this need is the \textit{counterfactual}:  a counterfactual for a given input and classifier is a similar input that results in a different classification.  Suppose a classifier is designed to determine whether a loan application represents a good or bad credit risk.  If the classifier determines a loan to be a bad credit risk, a counterfactual would be a modified loan application that is classified as a good credit risk, e.g. the individual in a loan application is a bad credit risk, but an otherwise identical applicant who is $5$ years younger with a \$$500$ higher monthly income would be a good credit risk.  \cite{original_CF} first suggested the use of \textit{Counterfactuals Explanations} (CE) to help understand classifiers' decisions making.  Subsequent works explored the use of counterfactuals to help individuals change an undesirable classification \cite{linearrecourse_CF,algorithmicrecourse_CF,face_CF}.  Returning to the example of an individual turned down for a loan, this type of counterfactual would not suggest an individual decrease their age (clearly impossible), but rather make practical changes such as pay off all credit card debt and request a $10\%$ smaller loan.  These counterfactuals came to be known as \textit{Actionable Counterfactuals} (AC) or \textit{Algorithmic Recourses} (AR).  Although, these counterfactuals change a classifier's decision, it can not be assumed they will have the same affect on the real world \cite{philosophy}, e.g. a change that causes a classifier to determine someone is a good credit risk may not increase the person's odds of paying off the loan in reality.  \cite{ICR_CF} point out that a counterfactual could change a classifiers decision without changing the real world if the modifications are not causally linked to the output.  For example, having a mailing address in an affluent neighborhood may correlate to higher odds of paying off a loan and changing the address could affect a classifiers decision, but there is no causal link.  Accordingly, telling an applicant to change their mailing address to a P.O. box in a wealthy neighborhood would not improve their chances of paying off a loan.  \cite{ICR_CF} proposed a framework to ensure modifications are causally linked to the output which they called \textit{Improvement-Focused Causal Recourse} (ICR).

\begin{figure*}[t]
\begin{center}
\includegraphics[width = 1\textwidth]{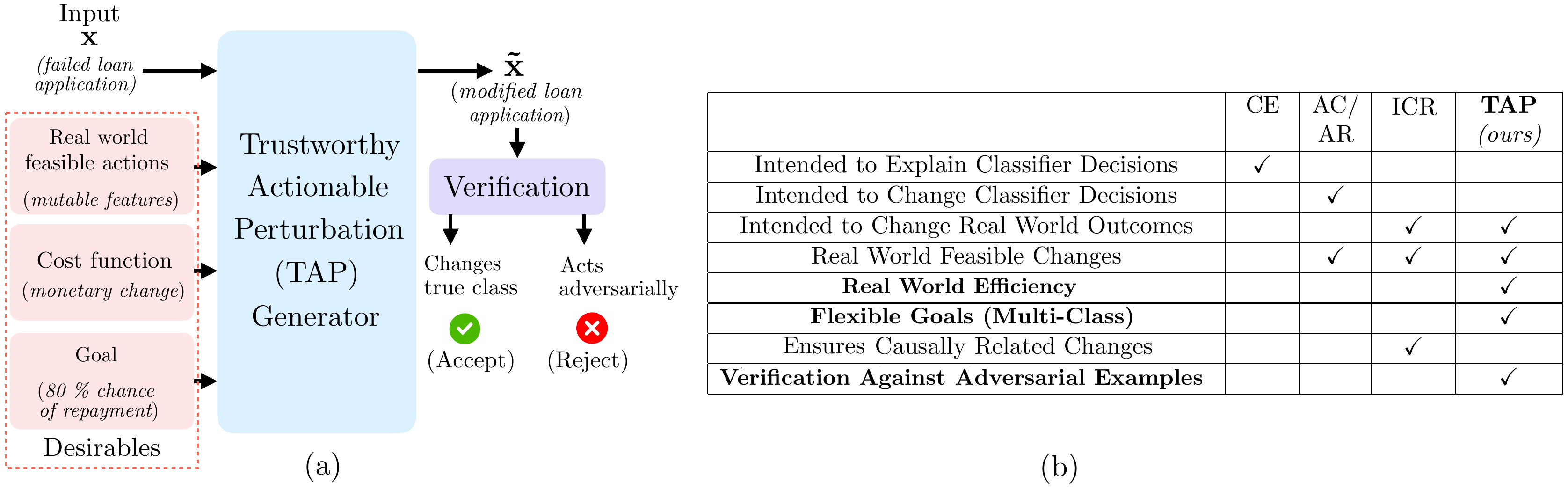}\label{img:outline}
\end{center}
\vspace{-13pt}
\caption{a) Overview of the framework for creating Trustworthy Actionable Perturbations (TAP). b) Comparison of  objectives and features of TAP and Counterfactual Explanations (CE) \cite{original_CF}, Actionable Counterfactuals/Algorithmic Recourse (AC/AR) \cite{linearrecourse_CF,algorithmicrecourse_CF,face_CF}, Improvement-Focused Causal Recourse (ICR) \cite{ICR_CF}.}\vspace{-10pt}
\end{figure*}

In this paper, we focus on tackling new challenges for this problem, which have not been addressed in prior work.
Trustworthy Actionable Perturbations (TAP) focus on three novel  improvements for affecting real world outcomes.

\underline{Trustworthiness Against Adversarial Examples}:  \cite{original_AA} showed that ML classifiers are brittle and small modifications to an input can cause misclassifications in otherwise accurate classifiers.  Among the various definitions of adversarial behavior, we use the definition: modifications to a data point are adversarial if they cause a classifier to be far less accurate on modified data points than the original data \cite{robustness_def}.  These modified inputs are called \textit{adversarial examples} and the algorithms that create them are called \textit{adversarial attacks}.  The algorithms that create counterfactuals are very similar to adversarial attacks and \cite{adversarial_couterfactual_analysis} showed they produce similar outputs, which leads to the troubling conclusion that many counterfactuals may act as adversarial examples and change the classifier decision (individual is now offered a loan) without changing the true underlying class probabilities (individual is still likely to default on the loan).  The adversarial vulnerability of classifiers is separate from causality concerns.  For this reason, we introduce a novel two step procedure where (1) we generate a suggested change and (2) we use an independently trained \textit{verifier} to certify that this change is not acting as an adversarial example.  We present a methodology for training this verifier and provide analytical results showing that it is PAC-learnable (Theorem \ref{theorem_PAC}).

\underline{Flexible Goal Definition}:  AC/AR focus solely on the final classification of a data point, but this may not always be  sufficient or feasible. 
For instance a valid AC/AR may lead to a $51\%$ likelihood of paying off a loan, but this may not satisfy the individual.  Additionally, a change that  improves a cancer patient's odds of survival  form $15\%$ to $40\%$ would not constitute a valid AC/AR even though it would be very useful.   
Accordingly, our framework defines goals through a \textit{target set} of acceptable outcomes that can be tailored to an individual's needs, and we demonstrate how these target sets can be designed.  We note that ICR \cite{ICR_CF} and one of the AC/AR methods 
\cite{multiobjective_CF}  propose the use of goals other than final classification, but our formulation is more flexible and applies to multi-class scenarios. We develop a principled measure of reward by defining a distance to the target set using statistical divergence. We analyze this distance theoretically in Theorem \ref{theorem_closedform}.

\underline{Real World Efficiency}:  Previous works on  CE and AC/AR reduce the amount of changes made by a counterfactual by minimizing a weighted $\ell$-norm of the changes (with the exception of \cite{actionsequence_CF}), however these norms often fail to represent the real world cost of making changes.  Alternatively, we minimize a cost measure built specifically to reflect real world costs of a change.  By using this measure of real world cost and principled measure of rewards (distance to target set), TAP can suggest more efficient advice. We present a few examples of the utility of producing efficient advice through TAP: (a) Suggest the course of treatment that would double a patient's odds of survival while requiring the least staff hours.  (b) List the skills an job applicant could acquire in the least amount of time that would lead to a high probability of receiving an interview.  (c) Find the cheapest modifications to a product that would bring it into a more premium price range and enhance marketability.  We illustrate through experiments on real world data how the use of application specific cost functions leads to more efficient advice.

\vspace{5pt}
Figure \ref{img:outline}(a) illustrates our framework of \textit{Trustworthy Actionable Perturbations} (TAP) for using feasible actions, true cost and an individualized goal to create an efficient change, which is then verified to ensure that the change affects the true class probabilities instead of acting adversarially.  Our goal to change the true class probabilities (real world outcomes) differs from previous CE and AC/AR works that seek only to change the classifier's decision.  We share our goal with ICR which is focused on ensuring that only features causally related to the class are modified.  Our framework, on the other hand, focuses on ensuring that the changes do not exploit the brittleness of ML classifiers and cause misclassifications.  This can occur regardless of whether modified features are causally related to the output.  Figure \ref{img:outline}(b) provides a summary of the objectives and features of various existing approaches alongside TAP.

\vspace{0pt}
\section{Trustworthy Actionable Perturbations}
\label{sec:definition}
\vspace{0pt}

\textbf{Problem Setting and Goals:} Suppose there is an unknown distribution $(\mathbf{x},C) \sim \mathcal{D}$.  Here $\mathbf{x}$ is a member of the input space  $\mathcal{X} \subset \mathbb{R}^d$  and $C \in \{1,...,k\}$ is the class of $\mathbf{x}$.  We define the true class probabilities $\mathbf{y}(\mathbf{x}) := \left (\mathbb{P}(C=1|\mathbf{x}),...,\mathbb{P}(C=k|\mathbf{x}) \right )$.  
We let $\mathcal{Y}$ denote the $k$-simplex and use a classifier $M:\mathcal{X}\rightarrow\mathcal{Y}$ to estimate $\mathbf{y}(\mathbf{x})$.  
Our goal in designing TAP is as follows:  \textit{Given an input $\mathbf{x}$ with an undesirable classification $M(\mathbf{x})$, find the most efficient real world actions to create a modified input $\tilde{\mathbf{x}}$ such that the corresponding true probabilities $\mathbf{y}(\tilde{\mathbf{x}})$ (and not just $M(\mathbf{\tilde{x}})$) are more desirable.}

\textbf{Real World Actionability:}  
TAP should only suggest modifications that are feasible in the real world (e.g., not decreasing an individual's age). To this end, we introduce: \textit{the Actionable Set $\mathcal{A}(\mathbf{x})$ of a data point $\mathbf{x}$ as the set of all perturbations of $\mathbf{x}$ that are feasible in the real world.}  For example, if $\mathcal{X}$ represents loan applications with $x_1$ the age of the applicant, $x_2$ the applicant's credit score, $x_3$ the amount of credit and $x_4$ the loan duration, the actionable set could be
$\mathcal{A}(\mathbf{x}) = \{ \tilde{\mathbf{x}}\in\mathcal{X} | \tilde{x}_1 = x_1, \tilde{x}_2 = x_2\}$,
i.e. the applicant can change the size and duration of the loan they request, but not their age or credit score.  Previous works have examined the complexities of actionability including  causal relations between inputs, e.g. one can't increase their education without an increase in age \cite{causalconstraint_CF,imperfectcausal_CF}.  All of these considerations, as well as a limiting changes to attributes which are believed to have a causal link to the output, can be incorporated into $\mathcal{A}(\mathbf{x})$.

\textbf{Efficiency:}  The definition of the most efficient change depends on the context of the problem and could involve a well defined value such as ``cost in dollars'' or more nebulous value such as ``amount of effort required.''  We characterize this value with a function $d_\mathcal{X}:\mathcal{X}\times\mathcal{X}\rightarrow \mathbb{R}$,  where $d_{\mathcal{X}}(\mathbf{x},\tilde{\mathbf{x}})$ is the cost of changing $\mathbf{x}$ to $\tilde{\mathbf{x}}$.  For example, if $\mathbf{x}$ and $\tilde{\mathbf{x}}$ represent resumes, then $d_{\mathcal{X}}(\mathbf{x},\tilde{\mathbf{x}})$ could represent the time it would take to acquire the attributes listed on resume $\tilde{\mathbf{x}}$, but not on $\mathbf{x}$.  We note this function may not be a true distance measure.  For example, if $d_{\mathcal{X}}$ represents the difference in financial cost between two courses of medical treatment, then $d_{\mathcal{X}}(\mathbf{x},\tilde{\mathbf{x}})$ should be negative when  $\tilde{\mathbf{x}}$ is more affordable than $\mathbf{x}$.  

\textbf{Desirability:} We now define what we mean by a desirable outcome\textemdash the goal of a TAP.  \textit{The Target Set $T$ is the set of all elements of $\mathcal{Y}$ that would be an acceptable result of a TAP.}
If we wish to belong to a desirable class $w$ with probability no less than $p$, the target set would have the form $T = \{{\mathbf{z}}\in \mathcal{Y} | z_w \geq p\}.$  If our goal is rather to avoid some undesirable class $u$, $T$ could be of the form $T = \{{\mathbf{z}}\in \mathcal{Y} | z_u \leq q\}$
for a fixed $q$.  
More generally, if we wish to belong to a set of desirable classes $\mathcal{W}$ with probability at least $p$ and we wish to belong to a set of undesirable classes $\mathcal{U}$  with probability no greater than $q$, we would use
\begin{equation}\label{gen_target_fam}
    T = \left \{\mathbf{z} \in \mathcal{Y} \bigg | \sum_{i\in\mathcal{W}}z_i \geq p ,  \sum_{i\in\mathcal{U}}z_i \leq q\right\}.
\end{equation}
We must quantify how close an TAP comes to achieving its goal in a principled manner.  To do this, we first choose a measure of statistical distance $D({\mathbf{y}} || \mathbf{z})$ (we use Kullback-Leibler (KL) Divergence). We then denote $d_{\mathcal{Y}}({\mathbf{y}},T)$ as the distance of ${\mathbf{y}}$ to the target set $T$, defined as follows:
\begin{align}\label{gen_distance_measure}
d_{\mathcal{Y}}({\mathbf{y}},T) := \inf_{\mathbf{z}\in T} D({\mathbf{y}} || \mathbf{z}).
\end{align}
We  may now formally define Trustworthy Actionable Perturbations.  Let $\epsilon$ represent budget \textemdash the amount of work we are willing to perform, and $\delta$ represent tolerance \textemdash how close the final result is to our target set $T$.

\begin{definition}[$(\epsilon,\delta)$-Trustworthy Actionable Perturbation]\label{def:ap}
$\tilde{\mathbf{x}}$ is an $(\epsilon,\delta)$-trustworthy actionable perturbation for $\mathbf{x}$ and a target set $T$ if
\begin{enumerate}
    \vspace{-4pt}
    \item $d_\mathcal{X}(\mathbf{x},\tilde{\mathbf{x}}) \leq \epsilon$
        \vspace{-4pt}
    \item $d_{\mathcal{Y}}({\mathbf{y}}(\tilde{\mathbf{x}}),T) \leq \delta$     \vspace{-4pt}
    \item $\tilde{\mathbf{x}}\in \mathcal{A}(\mathbf{x}).$
\end{enumerate}
\end{definition}
In order to verify the second condition we must be able to calculate $d_\mathcal{Y}$.  Fortunately, the optimization problem in (\ref{gen_distance_measure}) has a differentiable closed form solution when $D({\mathbf{y}} || \mathbf{z})$ is an $f$-divergence: a broad class of measures including KL-divergence, total-variation (TV) and other commonly used statistical distances.  An $f$-divergence is defined as $D({\mathbf{y}} || \mathbf{z}) = \sum_{i=1}^k \mathbf{z}_i f \left (\frac{{\mathbf{y}}_i}{\mathbf{z}_i} \right )$, where $f$ is a convex function satisfying $f(1)=0$ and $f(0) = \lim_{x\to 0^+}f(x)$ \citep{info_text}.  Theorem \ref{theorem_closedform} describes the solution to (\ref{gen_distance_measure}).

\begin{figure}[t]
\begin{center}
\includegraphics[width = 7.5cm]{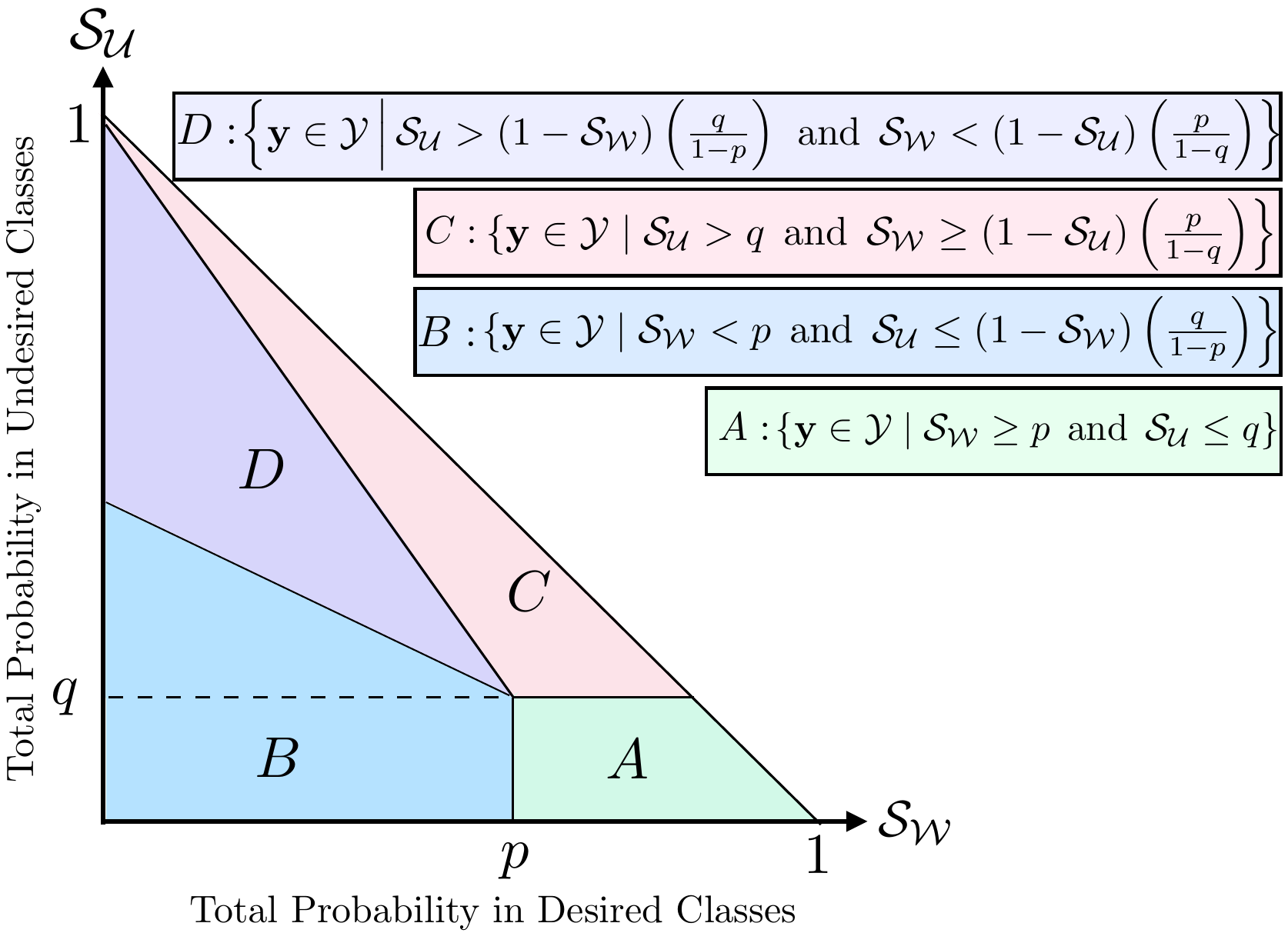}\label{img:probability_partitions}
\end{center}
\vspace{-10pt}
\caption{Illustration of the partition on $\mathcal{Y}$ used to calculate the distance from the target set $T$ in Theorem \ref{theorem_closedform}.  Although the cost function takes different functional form(s) in the four regions, it is continuously differentiable in the entire space.}
\vspace{-10pt}
\end{figure}

\begin{theorem}\label{theorem_closedform}
    If $D({\mathbf{y}} || \mathbf{z})$ is an $f$-Divergence with twice differentiable $f$ and $T$ is of form (\ref{gen_target_fam}), then 
\begin{equation}\label{closed_form}
d_{\mathcal{Y}}({\mathbf{y}},T)=
    \begin{cases}
        0 & \text{if } {\mathbf{y}}\in A\\
        
        pf\left( \frac{\mathcal{S}_\mathcal{W}}{p} \right ) + (1-p)f\left( \frac{1-\mathcal{S}_\mathcal{W}}{1-p} \right )& \text{if } {\mathbf{y}}\in B\\
        
        qf\left( \frac{\mathcal{S}_\mathcal{U}}{q} \right )+(1-q)f\left( \frac{1-\mathcal{S}_\mathcal{U}}{1-q} \right )& \text{if } {\mathbf{y}}\in C\\

        pf\left( \frac{\mathcal{S}_\mathcal{W}}{p} \right ) + qf\left( \frac{\mathcal{S}_\mathcal{U}}{q} \right )& \text{if } {\mathbf{y}}\in D \\
        +(1-p-q)f\left( \frac{1- \mathcal{S}_\mathcal{W}-\mathcal{S}_\mathcal{U}}{1-p-q} \right )
    \end{cases}
\end{equation}
where $ \mathcal{S}_\mathcal{W} = \sum _{i \in \mathcal{W}} {y}_i$, $\mathcal{S}_\mathcal{U} = \sum _{i \in \mathcal{U}} {y}_i$ and the sets $A,B,C$ and $D$ are a partition of $\mathcal{Y}$ defined and visualized in Figure \ref{img:probability_partitions}.
Furthermore, $d_{\mathcal{Y}}({\mathbf{y}},T)$ is continuously differentiable in $\mathbf{y}$ over its entire domain.
\end{theorem}

Equation (\ref{closed_form}) in Theorem \ref{theorem_closedform} is easily calculable and continuously differentiable despite its piece-wise form, which will be significant when creating TAP (see Section \ref{sec:creation}).  The proof of Theorem \ref{theorem_closedform} involves showing that optimization problem (\ref{gen_distance_measure}) is convex and finding a value $\mathbf{z}$ that satisfies the KKT conditions.  This proof and additional results about $d_\mathcal{Y}$ are found in the Appendix \ref{APP:stat_dist}.

\textbf{Real-world Verifiability of TAP}: Note that TAP are defined with respect to the true class probabilities ${\mathbf{y}}(\tilde{\mathbf{x}})$ because TAP should have an effect in the real world.   Notwithstanding, ${\mathbf{y}}(\tilde{\mathbf{x}})$ is unknown and we must use $M(\tilde{\mathbf{x}})$ to create our TAP (more details in Section \ref{sec:creation}), which introduces the risk that we might produce an $\tilde{\mathbf{x}}$ that has the desired effect on $M(\tilde{\mathbf{x}})$ but not ${\mathbf{y}}(\tilde{\mathbf{x}})$ (like an adversarial example).
This is of particular concern because TAP and all other counterfactuals are created by solving an optimization problem of the form 
 \begin{equation}
    \tilde{\mathbf{x}} = \arg\min_{\tilde{\mathbf{x}}}\hspace{0.1cm} loss(\tilde{\mathbf{x}},w)+\lambda\cdot dist(\tilde{\mathbf{x}},\mathbf{x})\label{gen_optimization},
\end{equation}
which is precisely how most adversarial examples are created \cite{adversarial_couterfactual_analysis}. 
When counterfactuals were first introduced to ML \cite{original_CF}, the concern that counterfactuals would act as adversarial examples was dismissed because the adversarial attacks of the time 1) modified many more features than counterfactuals and 2) were targeted almost exclusively at image data whereas counterfactuals were proposed for use on tabular data.  Since that time, \citet{hardness_robustness,onepixel_AA} demonstrated that adversarial attacks can be effective when changing a very small number of features,
 and several works  \citep{imperceptible_TAB,notequal_TAB,sony_TAB,mastercard_TAB} have shown that adversarial examples exist on tabular data sets.  This implies that verification is necessary to achieve results that can be trusted to change the true class probabilities.

Verifying $\tilde{\mathbf{x}}$ may appear similar to detecting adversarial examples, which  has been the object of significant research \citep{MLLOO_ADD,oddodds_ADD,SHAPdetect_ADD,carlinidetection_ADD} 
with no satisfactory solution.
Fortunately, we have an important advantage over detecting adversarial examples: \textit{we know the original data point $\mathbf{x}$ and exactly how it was modified, i.e., $\tilde{\mathbf{x}}$}.  To capitalize on this knowledge, we propose a novel verification procedure using a classifier $V:\mathcal{X}\times\mathcal{X}\rightarrow [0,1]$ which compares two inputs simultaneously and predicts the probability of the inputs belonging to the same class: the value of $V(\mathbf{x},\tilde{\mathbf{x}})$ estimates $\mathbb{P}(C=\tilde{C}|\mathbf{x},\tilde{\mathbf{x}})$.  
Because $V$ has a different classification task from $M$, attacks targeted against $M$ should not be effective against $V$, and we can use the discrepancy between estimates of $M$ and $V$ to determine if an $\tilde{\mathbf{x}}$ acts adversarially on $M$.  In order to make this comparison, we use the fact that $\mathbb{P}(C=\tilde{C}|\mathbf{x},\tilde{\mathbf{x}})$ can also be estimated using $M$ by calculating $\sum_{i=1}^k M_i(\mathbf{x})M_i(\tilde{\mathbf{x}})$.  If $\tilde{\mathbf{x}}$ acts adversarially we would expect $\sum_{i=1}^k M_i(\mathbf{x})M_i(\tilde{\mathbf{x}})$ to be very small while $V(\mathbf{x},\tilde{\mathbf{x}})$ is large.  If $\tilde{\mathbf{x}}$ is not adversarial we would expect similar values from both $\sum_{i=1}^k M_i(\mathbf{x})M_i(\tilde{\mathbf{x}})$ and $V(\mathbf{x},\tilde{\mathbf{x}})$.   Accordingly, we define
\begin{equation}\label{eqn:discrepancy}
\Delta (\mathbf{x},\tilde{\mathbf{x}}) := \left |V(\mathbf{x},\tilde{\mathbf{x}}) - \sum_{i=1}^k M_i(\mathbf{x})M_i(\tilde{\mathbf{x}})\right |,
\end{equation}
 and verify that an $\tilde{\mathbf{x}}$ is trustworthy only if $\Delta (\mathbf{x},\tilde{\mathbf{x}})<\gamma$.
 In Section \ref{sec:creation}, we describe how we selected the threshold $\gamma$.

\textbf{Training a Verifier \& PAC Learnability}:  In order to create $V$, we must have data on which it can be trained.  We build this \textit{difference training data} by creating all possible pairs of elements from our original training data and labeling the pairs by whether they belong to the same class ($1$ for the same class, $0$ for different classes).  If the original training data is $\{\mathbf{x}^{(i)},C^{(i)}\}_{i=1}^n$, the difference training data is $\{(\mathbf{x}^{(i)},\mathbf{x}^{(j)}), z^{(i,j)}\}_{1\leq i,j\leq n}$, where $z^{(i,j)}=\mathbbm{1}[C^{(i)} = C^{(j)}]$. We use the same architecture for $V$ as $M$ (only changing the number of inputs and outputs), but differing architectures could also be used.  Now that we have a method for training $V$, we show that training in this way leads to a generalizable verifier. To this end, we next present a probably approximately correct (PAC) bound on $V$'s generalization gap which depends on $n$ (number of training samples), $k$ (number of classes), and $d$ (data dimensionality).  

\begin{theorem}\label{theorem_PAC}
   Let $\mathrm{R}(V)$ be the true risk of a verifier $V$ over data drawn from $\mathcal{D}$ and $\hat{\mathrm{R}}_\mathcal{S}(V)$ be the empirical risk over a sample $\mathcal{S}$ of labelled point pairs drawn i.i.d. from $\mathcal{D}$.  Both risks are defined using a bounded loss function $\ell(\cdot,\cdot)\leq B_\ell$.  Also let $V$ be selected from a function class $\mathcal{V}$.  Then for any $\delta\in(0,1)$, with probability $(1-\delta)$, the following bound on the generalization gap holds.
   \begin{equation}
       \sup_{V\in\mathcal{V}}\left | \mathrm{R}(V)- \hat{\mathrm{R}}_S(V) \right | =  \mathcal{O}\left ( \left (\frac{k}{\sqrt{n^2-k^2n}} \right )^{1/d} \right ) \label{eq:PAC}
   \end{equation}
   Here the terms with explicit dependence on $\delta$ have been suppressed because they are dominated by the term in (\ref{eq:PAC}).  The precise generalization bound is presented in \eqref{eq: final-PAC-bound} in the Appendix.
\end{theorem}

To prove Theorem \ref{theorem_PAC}, we construct a definition of risk that fits this new learning scenario (i.e., learning if two samples are from the same class or not, as opposed to conventional classification).  This risk takes into account that we expect large imbalances between the number of point pairs from the same class and from different classes. In order to obtain the bounds on the generalization gap, we expand this risk into a sum of terms which can be bounded with existing Rademacher complexity PAC-methods.  Finally, we bound the growth of these Rademacher complexity terms as a function of $n, k$ and $d$ to arrive at (\ref{eq:PAC}).  
The complete proof, including detailed definitions of $\mathrm{R}(V)$ and $\hat{\mathrm{R}}_S(V)$ as well as additional discussion, is presented in the Appendix \ref{APP_PAC}.

\vspace{5pt}
\begin{remark}
The bound in Theorem \ref{theorem_PAC} is small as long as $n\gg k^2$ and $n$ is exponentially larger than $d$.  The relation between $n$ and $k$ is crucial because it implies that the denominator $\sqrt{n^2-k^2n}\approx n^2 \gg k$.  This differs from typical PAC bounds where the primary requirement is $n$ be exponentially larger than $d$ (Theorem 4.3 in \cite{gottlieb2016adaptive}) and have mild dependence on the number of classes $k$.  The key implication of this result is: \textit{when using a verifier as described in this paper, as the data sets used increase in number of classes $k$, it is essential that the amount of training data increases at a rate of $\sqrt{k}$}.


\end{remark}

\vspace{-6pt}
\section{Generating TAP}\label{sec:creation}

\textbf{Two Step Creation Method:} We now present and discuss the general optimization framework for creating TAP. Ideally, we would like to solve the following optimization problem: $\arg\min_{\tilde{\mathbf{x}} \in \mathcal{A}(\mathbf{x})} d_{\mathcal{Y}} (\tilde{\mathbf{y}},T) + \lambda d_\mathcal{X}(\tilde{\mathbf{x}},\mathbf{x})$, 
where the scalar parameter $\lambda$ balances the effort($\epsilon$)-reward($\delta$) trade-off.  
Solving this optimization would be guaranteed to create an effective TAP; unfortunately ${\mathbf{y}}(\tilde{\mathbf{x}})$ is unknown and we cannot solve this problem directly. 
Instead propose the following two-step procedure where: in Step $1$, we treat $M(\tilde{\mathbf{x}})$ as a surrogate for ${\mathbf{y}}(\tilde{\mathbf{x}})$, and in Step $2$, we use a verification algorithm to ensure that $\tilde{\mathbf{x}}$ is not just fooling the classifier.  
\begin{gather}
    \textbf{Step 1 :}\arg\min_{\tilde{\mathbf{x}} \in \mathcal{A}(\mathbf{x})} d_{\mathcal{Y}} (M(\tilde{\mathbf{x}}),T) + \lambda d_\mathcal{X}(\tilde{\mathbf{x}},\mathbf{x})\label{eqn:our_opt}\\
    \downarrow\notag\\
    \textbf{Step 2:}\text{ Verify } M(\tilde{\mathbf{x}})\approx {\mathbf{y}}(\tilde{\mathbf{x}})\text{  i.e. } \Delta(\mathbf{x},\tilde{\mathbf{x}})\leq\gamma\\
    \downarrow\notag\\
    \text{TAP}\notag
    \label{optimization_problem}
\end{gather}

\textbf{Solving Step 1:} We solve (\ref{eqn:our_opt}) using gradient descent which requires us to use differentiable models $M$ and formulate $d_\mathcal{X}$ in a differentiable manner ($d_\mathcal{Y}$ is differentiable according to Theorem \ref{theorem_closedform}).  We modify our gradient descent to address two challenges.  (1) We must insure that our solution is actionable: $\tilde{\mathbf{x}} \in \mathcal{A}(\mathbf{x})$.  (2) Our solution $\tilde{\mathbf{x}}$ must follow any formatting rules associated with the data set (for instance, Boolean variables must be either 0 or 1, categorical features must respect one-hot encoding, etc.).  A perturbation that follows these formatting rules is called \textit{coherent}.  
To solve these two difficulties, we first assume $\mathcal{A}(\mathbf{x}) = \{\tilde{\mathbf{x}} | l_i \leq \tilde{\mathbf{x}}_i \leq u_i,  1\leq i\leq d\}$
for some set of lower bounds $\{l_i\}_{i=1}^d$ and upper bounds $\{u_i\}_{i=1}^d$.  An attribute is \textit{immutable} if $l_i = u_i$.
We ensure actionability by setting all elements of the gradient corresponding to immutable features to zero and adding a large  penalty $b(\tilde{\mathbf{x}})$ term to the objective function which punishes points for leaving the actionable set.  
To ensure coherence, we project the result of our gradient descent onto the coherent space by using a function $cond:\mathbb{R}^m\rightarrow\mathcal{X}$ which performs the appropriate value rounding to make an input coherent.  We found it useful to introduce a second penalty term $p(\tilde{\mathbf{x}})$ which requires that any one-hot encoded features sum to $1$.  This ensures our answers never stray too far from a coherent point and improves robustness.  Details on $b$, $p$ and $cond$ are found in the Appendix \ref{APP:objective}.  In practice we also found it useful to replace regular gradient descent with the ADAM algorithm \citep{ADAM}.

\begin{algorithm}[t]
\caption{Generating TAP}\label{alg:ap_gen}
\begin{algorithmic}
\STATE {\bfseries Input:} Classifiers $M$ \& $V$, point $\mathbf{x}$, target family $T$, learning rate $\alpha$, verification-cut off $\gamma$

\STATE $\tilde{\mathbf{x}}$ $\gets$ $\mathbf{x}$
\WHILE{$\tilde{\mathbf{x}}$ not converged}    
    \STATE $\mathbf{g} \gets \nabla_{\tilde{\mathbf{x}}} \left (d_{\mathcal{Y}} (M(\tilde{\mathbf{x}}),T) + \lambda d_\mathcal{X}(\tilde{\mathbf{x}},\mathbf{x}) + b(\tilde{\mathbf{x}}) + p(\tilde{\mathbf{x}}) \right )$

        \STATE $\mathbf{g}_j \gets 0$ for all immutable features $j$.

    \STATE $\tilde{\mathbf{x}}$ $\gets$ $\tilde{\mathbf{x}} - \alpha g$
\ENDWHILE
\STATE $\tilde{\mathbf{x}} = cond(\tilde{\mathbf{x}})$ (project onto the coherent space)
\STATE $\epsilon,\delta =d_\mathcal{X}(\tilde{\mathbf{x}},\mathbf{x}),d_{\mathcal{Y}}(M(\tilde{\mathbf{x}}),T)$

\IF{$\epsilon$ and $\delta$ requirements NOT met}
\STATE Adjust $\lambda$ (see text for explanation)
\STATE Return to while loop
\ENDIF
\IF{$\left |V(\mathbf{x},\tilde{\mathbf{x}}) - \sum_{i=1}^k M_i(\mathbf{x})M_i(\tilde{\mathbf{x}})\right | \geq \gamma$}
\STATE Adjust problem parameters (see text for explanation)
\STATE Restart algorithm
\ENDIF
\STATE return $\tilde{\mathbf{x}}$
\end{algorithmic}
\end{algorithm}

\begin{figure*}[t]
\begin{center}
\includegraphics[width = 1\textwidth]{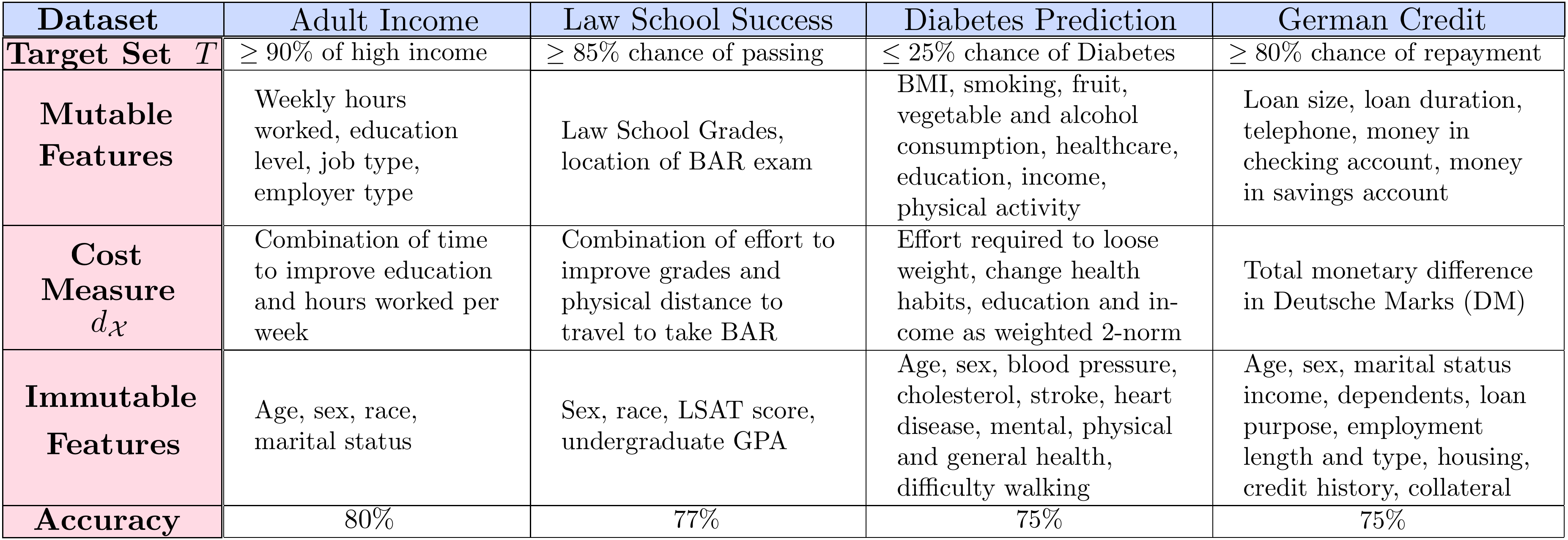}
\end{center}
\vspace{-10pt}
\caption{Table containing details on data sets used for testing.}
\vspace{-10pt}
\label{tab:data_summary}
\end{figure*}

\textbf{Solving Step 2:} In Section \ref{sec:definition}, we discussed the necessity of verification and suggested that an TAP can be trusted if $\Delta (\mathbf{x},\tilde{\mathbf{x}}) = \left |V(\mathbf{x},\tilde{\mathbf{x}}) - \sum_{i=1}^k M_i(\mathbf{x})M_i(\tilde{\mathbf{x}})\right |$ is smaller than a threshold $\gamma$.  Our process for choosing $\gamma$ starts with deciding on an acceptable risk of eliminating a truly effective TAP (we use 10\%). To find the $\gamma$ corresponding to this risk, we calculate $\Delta(\mathbf{x}^{(i)},\mathbf{x}^{(j)})$ for a sufficiently large number of pairs $(\mathbf{x}^{(i)},\mathbf{x}^{(j)})$ from the testing data such that $C^{(i)} \neq C^{(j)}$.  Finally, we pick $\gamma$ such that only the desired percentage of $\Delta(\mathbf{x}^{(i)},\mathbf{x}^{(j)})$ values (e.g. 10\%) are above $\gamma$.  The verification procedure is now reduced to eliminating any $\tilde{\mathbf{x}}$ that results in $\Delta (\mathbf{x},\tilde{\mathbf{x}})>\gamma$.

\textbf{Adjusting for Suitability and Verifiability:}  When creating TAP we will often have a particular budget ($\epsilon$) or tolerance ($\delta$) bound we need to satisfy.  To find a suitable TAP we repeat Step 1 of our process adjusting $\lambda$ until the desired budget or tolerance is met: increasing $\lambda$ to decrease $\epsilon$ and decreasing $\lambda$ to decrease $\delta$.  
It may also be appropriate to use a variety of $\lambda$ values and plot the $\epsilon$ and $\delta$ values of each resulting TAP (see Figure \ref{img:ac_plot}).  The user may then select a TAP they see as offering particularly good value.  When a TAP fails the verification step, there are a few recourses.  (1) Sometimes it is sufficient to decrease $\lambda$, putting greater emphasis on reaching the target set.
(2) ``Shrink'' the target set (increase the value of $p$ and decrease the value of $q$) in order to force the algorithm to find more effective changes.  (3)  Add a random perturbation to $\mathbf{x}$ in order to move the starting point away from the adversarial example.  The entire procedure is described in Algorithm \ref{alg:ap_gen}.

\vspace{-8pt}
\section{Experimental Results}
\label{sec:results}
\vspace{-5pt}

\textbf{Data Sets:} We compare TAP, counterfactuals and adversarial attacks on four data sets from different fields; data set details are found in Figure \ref{tab:data_summary} and the Appendix \ref{APP_datasets}.

\begin{figure*}[t]
\begin{center}
\includegraphics[width = 1\textwidth]{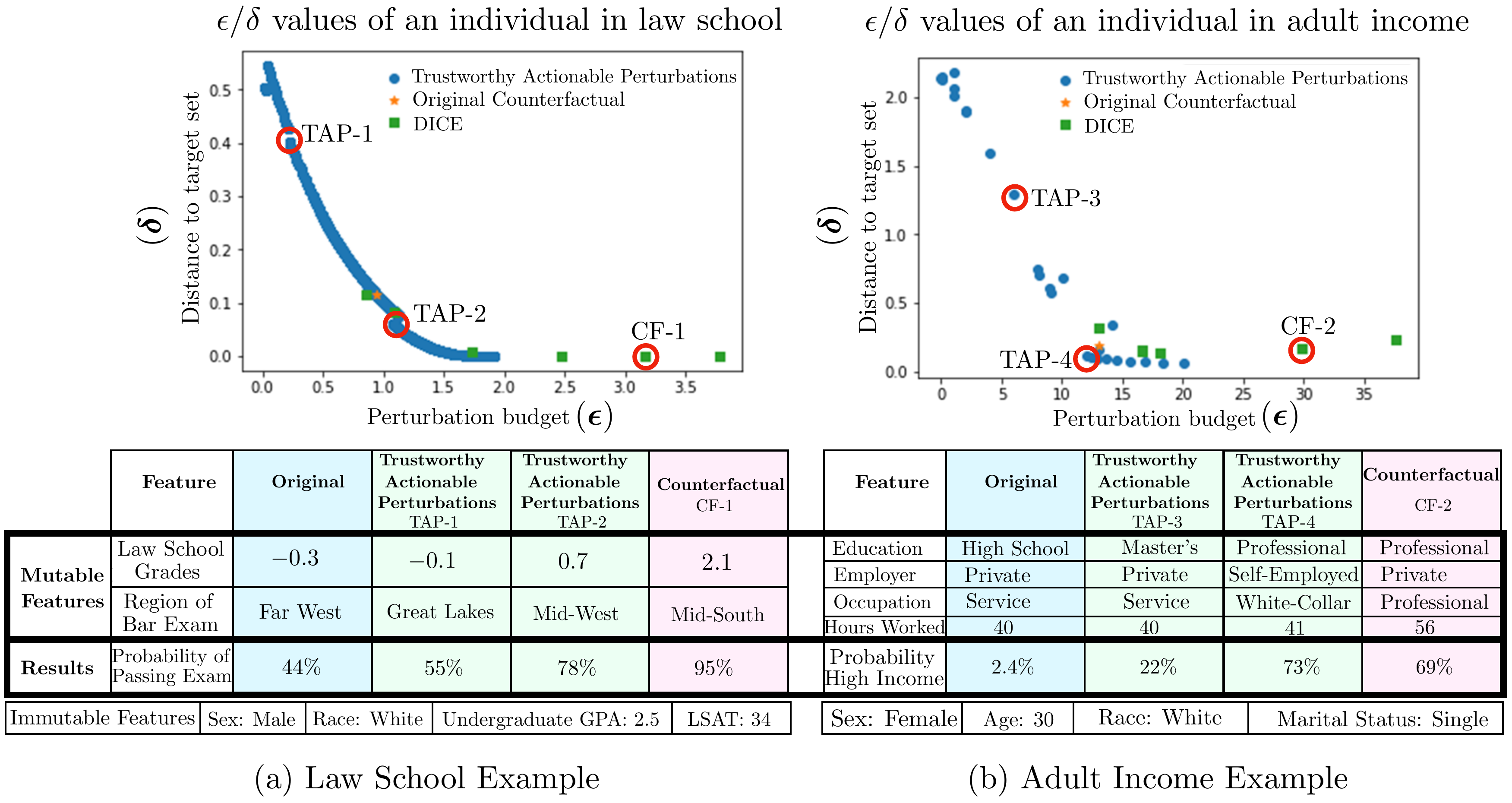}
\end{center}
\vspace{-10pt}
\caption{Cost-Benefit plots of TAP and counterfactuals for an individually from the Law School data set with grades measured in standard deviations from the mean (a) and an individual in the Adult Income data set (b).}\label{img:ac_plot}
\vspace{-10pt}
\end{figure*}

\underline{\textit{Adult Income}} \citep{adult_DS}: This data set contains demographic information on Americans labelled by whether they had a high income.  The actionable set $\mathcal{A}(\mathbf{x)}$ allows individuals to increase their education, change jobs and adjust their weekly work hours.  The cost function $d_\mathcal{X}$ sums the expected number of years to improve education, a one-year cost to change jobs and the square of the change in hours worked (weighted so an additional 3 hours of work per week is equal to a year spent on education).
\newline
\underline{\textit{Law School Success}} \citep{law_DS}: This data set contains information on law school students labelled by whether they passed the BAR exam.  $\mathcal{A}(\mathbf{x)}$ allows changes to law school grades (through more studying) and the region where the exam is taken. The cost function $d_\mathcal{X}$ sums the increase in grades and the physical distance travelled to take the BAR.  Moving to an adjacent region (Far West to North West) is weighted equal to increasing grades one standard deviation.
\newline
\underline{\textit{Diabetes Prediction}} \citep{diabetes_DS}: The individuals in this data set are labelled by whether they have diabetes. We define $\mathcal{A}(\mathbf{x})$ to allow changes in health habits, BMI, education and income.  We use a weighted 2-norm for $d_\mathcal{X}$ to represent the relative difficulty of making changes.  For example, starting to get regular physical activity is weighted the same as dropping one BMI.
\newline
\underline{\textit{German Credit}} \citep{germancredit_DS}: This data set contains loan applications.  In $\mathcal{A}(\mathbf{x})$, we allow for changes to the loan duration and size and funds in the checking and savings accounts.  We use  $d_\mathcal{X}$ to measure the total difference in Deutsche Marks (DM) over all elements of the application.   

\textbf{Other Methods:}  We compare our results against counterfactuals created using the original method proposed to create counterfactuals \citep{original_CF} and the diverse counterfactuals (DICE)  method in \citep{Diverse_CF}, the most cited methods in the literature.   These methods use an $\ell_p$ norm based cost function that often fails to reflect real world costs (see examples on the next page).  We also compare TAP against  the \citet{carliniwagner_AA} $\ell_2$ adversarial attack, one of the most well known and effective adversarial attacks.  The counterfactuals belong to the same actionable set as the TAP, but the adversarial examples are not limited to an actionable set and may not be coherent.

\textbf{Models:} Gradient boosted tree algorithms \cite{GBT} are considered state of the art architectures for tabular data classification \cite{tree_superior}.  Unfortunately, these models are not differentiable and cannot be used with our framework. Instead we use neural networks which we tuned until they provide accuracy on par with gradient boosted tree models on the same data set.  Details on our models' structure, training are given in Appendix \ref{APP:models}.

\begin{figure*}[t]
\begin{center}
\includegraphics[width = 1\textwidth]{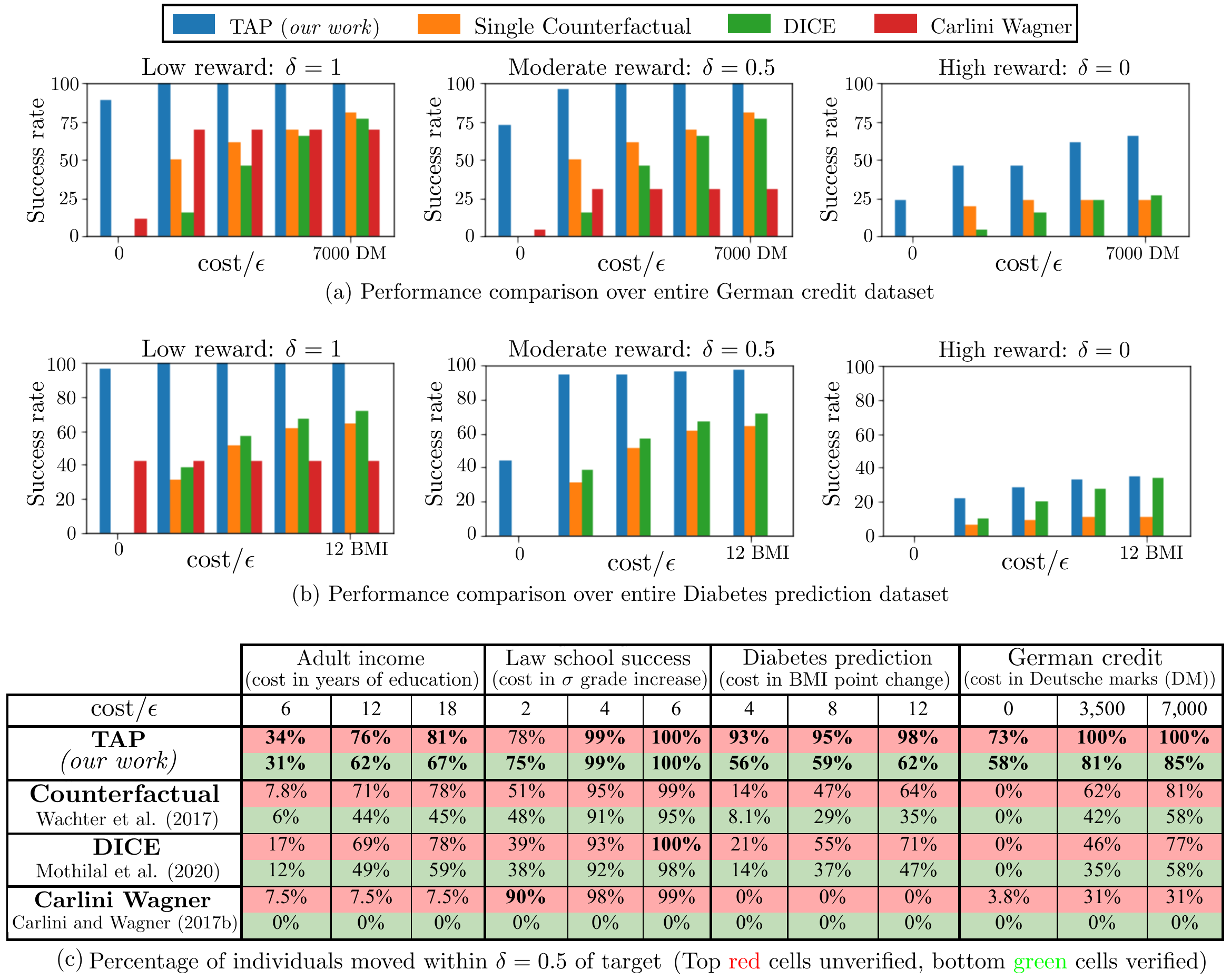}
\vspace{-20pt}
\caption{a) \& b) show average success rate for moving individuals within a variety of distances ($\delta$) to the target set.  The y-axis shows the percentage of individuals  within the goal distance, and the x-axis, represents different costs ($\epsilon$ values).  c) Summarizes success values for all data sets.  The upper {\color{red}(red)} value for each row is the success rate before the verification procedure and the lower {\color{green}(green)} value is the success rate after verification with a $10\%$ chance of rejecting valid examples.}
\label{fig:full_comparison}
\vspace{-20pt}
\end{center}
\end{figure*}

\textbf{Representative Examples of TAP and Trade-off between cost/desirability:}  We first examine two representative examples of how TAP behave differently than counterfactuals for specific individuals.
Figure \ref{img:ac_plot} shows a plot of the $\epsilon$/$\delta$ values of TAP and counterfactuals for one individual in the Law School data set and one individual in the Adult Income data set.  We examine the results from the Law School data set: The TAP labelled TAP-1 suggests only a mild ($0.2$ standard deviation) increase in grades and the relatively short move from the Far West to the Great Lakes region resulting in a small $11\%$ increase in the chance of passing the BAR.  On the other hand, TAP-2 suggest a larger increase in grades and a longer move which results in a much larger $34\%$ increase to the odds of success.  Finally the counterfactual CF-1 suggest an enormous increase in grades and massive cross country move to achieve $51\%$ increase in the odds of success.  Turning our attention to the Adult Income example: TAP-3 suggests a relatively simple increase in education to the masters level resulting in a $20\%$ increase to the odds of a high income.  Alternatively, TAP-4 achieves an $71\%$ increase by suggesting far more changes including a professional degree and becoming self-employed.  The counterfactual CF-2 does not suggest becoming self-employed and produces a smaller $67\%$ increase in the odds of high income despite also suggesting a professional degree and a drastic $16$ hour increase in the hours worked per week.  

These examples illustrates two trends: 1) TAP offer both low-cost/low-reward (large-$\delta$/small-$\epsilon$)  and high-cost/high-reward options, whereas counterfactual methods \cite{original_CF,Diverse_CF} only offer high-cost options. This is because TAP are defined by distance to the target set, but counterfactuals are defined as belonging to the desirable class.  That rules out any advice that doesn't result in the desirable class being the most likely class.  2) Counterfactuals are prone to suggesting very high-cost outliers. This has two main causes: (a) The $\ell_1$ norm used to create the counterfactuals does not accurately represent real world effort.  For example this norm considers any move in region to cost the same regardless of actual distance.
(b) Because counterfactuals do not use a target set, they are prone to  ``overshooting'' the desired goal.  For example $CF_1$ resulted in a $95\%$ chance of passing the BAR when our goal was only $85\%$.

\textbf{Comparison of TAP vs. Other Approaches:} We now compare TAP, counterfactuals \cite{original_CF,Diverse_CF} and CW attacks \cite{carliniwagner_AA} over the entire data sets. In Figure \ref{fig:full_comparison}:  Each bar chart refers to a particular data set and desired distance $\delta$ to the target set $T$.  Each bar shows the percentage of individuals that a method was able to move inside the goal $\delta$ at a variety of costs $\epsilon$.
(Bar charts for all data sets are found in the Appendix \ref{APP:results}.)
The table summarizes this information for all data sets with the upper {\color{red}(red)} value in each cell representing the data before the verification procedure and the lower {\color{green}(green)} value the success rate after the verification procedure.  Consider the bar chart on the top middle which refers to the German Credit data and a goal of $\delta = 0.5$ from the target (the same information as the last three columns of the table).  At a $\epsilon=0$ Deutsche Marks (DM) cost, TAP are able to move $73\%$ of individuals within the goal range by closing empty accounts.  Counterfactuals do not match this success until the cost $\epsilon = 7,000\text{DM}$, and CW attacks never achieve more than a $31\%$ success rate.  TAP outperform counterfactuals in all of the test scenarios.

\textbf{Impact and Effectiveness of Verifier:} The first important take away from the success rates after verification is that the verifier was 100\% effective at eliminating Carlini Wagner adversarial examples (visible in the bottom row of the table in Figure \ref{fig:full_comparison} c), implying that the verification method does indeed eliminate inputs that fool the classifier.  Importantly, the verification procedure also removes a significant number of TAP and counterfactuals.  Consider the second column of Figure \ref{fig:full_comparison} c:  Out of all TAP generated $14\%$ appeared effective but were eliminated by the verification procedure. Counterfactual methods fared even worse with $20\%$ to $27\%$ of counterfactuals eliminated.  This  reinforces the necessity of a verification procedure. 

\textbf{Concluding Remarks \& Future Work:} In this work, we proposed Trustworthy Actionable Perturbations (TAP) which leverage ML classifiers to find efficient actions to achieve real world results.  Our proposed framework introduces a novel verification procedure, flexible definition of goals, and principled reward measure for use in generating counterfactuals.
We demonstrated their effectiveness when compared to other methods on data sets from multiple fields.  Finally we note that our framework is flexible enough to incorporate contributions from previous works on counterfactuals such as individualized cost measures \cite{personalized_CF}, causal relations between  inputs \cite{causalconstraint_CF,imperfectcausal_CF}, causal relationships to the output \cite{ICR_CF}, and advanced optimization methods  \cite{rulebased_CF,mace_CF}.

\section*{Impact Statement}

As the use of AI and ML expands into critical applications such as healthcare, criminal justice, and hiring, the importance of explaining decisions deemed unfavorable and providing recourse to such users has grown significantly. In this context, our paper introduces a novel contribution aimed at making recourse mechanisms more trustworthy. We present a flexible framework, Trustworthy Actionable Perturbations (TAP), designed to generate cost-effective recourse which can ensure that the recourse being provided to users results in real-world changes. TAP can be useful to both end-users and institutions that suggest the recourse. The technical tools and the analytical results developed in the paper (including a flexible target set, and a novel pair-wise verification procedure) can also find use and lead to new insights for other problems such as cost-sensitive learning and adversarial defense.  

\vspace{-0.25cm}
\section*{Acknowledgements}

We thank the anonymous ICML reviewers and the area chairs for their insightful suggestions. 
This work was supported by NSF grants CAREER 1651492, CCF-2100013, CNS-2209951, CNS-1822071, CNS-2317192, and by the U.S. Department of Energy, Office of Science, Office of Advanced Scientific Computing under Award Number DE-SC-ERKJ422, and NIH Award R01-CA261457-01A1.


\bibliography{references}
\bibliographystyle{icml2024}

\newpage
\appendix
\onecolumn
\section{Appendix}
The Appendix is organized as follows:

\ref{APP:stat_dist} Proof of Theorem \ref{theorem_closedform} (Analysis of statistical distance $d_\mathcal{Y}$ to the target set $T$)\newline
\ref{APP_PAC} Proofs of Theorem \ref{theorem_PAC} (PAC generalization bounds for Verifier)\newline
\ref{APP_implementation}  Additional details about the implementation of experiments\newline
\ref{APP_datasets} Details about data sets and their corresponding cost functions\newline
\ref{APP:models}  Details about the models used\newline
\ref{APP:objective}  Details about the objective function used for optimization\newline
\ref{APP:results} Additional experimental results showing the comparative performance of TAP vs. other methods. 
\subsection{Proof of Theorem \ref{theorem_closedform} (Analysis of statistical distance $d_\mathcal{Y}$ to the target set $T$)}\label{APP:stat_dist}

Recall that our target sets have the form 
\begin{equation*}
    T = \left \{ \mathbf{z} \in \mathcal{Y} \left | \sum_{i\in\mathcal{W}}z_i \geq p \right. ,  \sum_{i\in\mathcal{U}}z_i \leq q\right\},
\end{equation*} where either $\mathcal{W}$ or $\mathcal{U}$ could be empty.
Also recall
\begin{align}
d_{\mathcal{Y}}(\tilde{\mathbf{y}},T) &= \min_{\mathbf{z}\in T} D_f (\tilde{\mathbf{y}}||\mathbf{z})= \min_{\mathbf{z}\in T} \sum_{i=1}^k{z}_i f\left  (\frac{\tilde{y}_i}{z_i}\right )\label{prob_dist}.
\end{align}
We must prove three facts: 1) $d_{\mathcal{Y}}(\tilde{\mathbf{y}},T)$ has the closed form found in equation (\ref{closed_form}), 2) This function is continuous, 3) the derivative of the function is continuous.  We begin by proving the closed form equation.

Our proof will be made easier by introducing notation $\mathcal{N} = (\mathcal{W}\cup\mathcal{U})^C$ as the neutral classes that are neither desirable nor undesirable.
We will use the fact that $1 = \mathcal{S}_\mathcal{W} + \mathcal{S}_\mathcal{U} + \mathcal{S}_\mathcal{N}$ to rewrite \eqref{closed_form} as

\begin{equation*}
d_{\mathcal{Y}}(\tilde{\mathbf{y}},T)=
    \begin{cases}
        0 & \text{if } \mathcal{S}_\mathcal{W}\geq p \text{ and } \mathcal{S}_\mathcal{U} \leq q\\
        
        pf\left( \frac{\mathcal{S}_\mathcal{W}}{p} \right ) + (1-p)f\left( \frac{\mathcal{S}_\mathcal{U}+\mathcal{S}_\mathcal{N}}{1-p} \right )& \text{if } \mathcal{S}_\mathcal{W} < p \text{ and } \mathcal{S}_\mathcal{U}\leq (1-\mathcal{S}_\mathcal{W})\left ( \frac{q}{1-p} \right)\\
        
        qf\left( \frac{\mathcal{S}_\mathcal{U}}{q} \right )+(1-q)f\left( \frac{\mathcal{S}_\mathcal{W}+\mathcal{S}_\mathcal{N}}{1-q} \right )& \text{if } \mathcal{S}_\mathcal{U} > q \text{ and } \mathcal{S}_\mathcal{W} \geq (1-\mathcal{S}_\mathcal{U})\left ( \frac{p}{1-q} \right)\\

        pf\left( \frac{\mathcal{S}_\mathcal{W}}{p} \right ) + qf\left( \frac{\mathcal{S}_\mathcal{U}}{q} \right )+(1-p-q)f\left( \frac{\mathcal{S}_\mathcal{N}}{1-p-q} \right )& \text{if } \mathcal{S}_\mathcal{U}>(1-\mathcal{S}_\mathcal{W})\left ( \frac{q}{1-p} \right) \\
        &\text{ and } \mathcal{S}_\mathcal{W}<(1-\mathcal{S}_\mathcal{U})\left ( \frac{p}{1-q} \right)
    \end{cases},
\end{equation*}
where $\mathcal{S}_\mathcal{W} = \sum_{i\in\mathcal{W}}\tilde{y}_i$, $\mathcal{S}_\mathcal{U} = \sum_{i\in\mathcal{U}}\tilde{y}_i$ and $\mathcal{S}_\mathcal{N} = \sum_{i\in\mathcal{N}}\tilde{y}_i$. 

The case where $\tilde{\mathbf{y}}\in T$ is obvious so we consider only the case where $\tilde{\mathbf{y}}\notin T$,  First note that $f$-divergence $D_{f}(\tilde{\mathbf{y}} || \mathbf{z})$ is convex in $\mathbf{z}$.  Furthermore $T$ is a convex set.  Therefore any $\mathbf{z}$ satisfying the KKT conditions is a minimizer.  The KKT conditions for this problem can be written as
\begin{align}
 \nabla \mathcal{L}(\mathbf{z}) = \vec{0}\label{gradarg}\\
 \sum_{i=1}^k z_i  = 1\label{prob_cond} \\
 p - \sum_{i \in \mathcal{W}}{z}_i \leq 0 \label{wanted_cond}\\
 \sum_{i \in \mathcal{U}}{z}_i - q \leq 0 \label{unwanted_cond}\\
 \mu_1, \mu_2 \geq 0 \label{d_feasibility} \\
 \mu_1 \left (p - \sum_{i \in \mathcal{W}}{z}_i \right ) = 0\label{comp_wanted}\\
 \mu_2 \left (q - \sum_{i \in \mathcal{U}}{z}_i \right ) = 0,\label{comp_unwanted}
 \end{align}
 where the Lagrangian  is defined by
$$
\mathcal{L}(\mathbf{z}) = \sum_{i=1}^k{z}_i f\left  (\frac{\tilde{y}_i}{z_i}\right ) + \lambda\sum_{i=1}^k z_i + \mu_1 \left (p - \sum_{i \in \mathcal{W}}{z}_i \right ) + \mu_2 \left (\sum_{i \in \mathcal{U}}{z}_i - q \right ).
$$
Note that we have neglected to explicitly state the requirement that $0 \leq z_i \leq 1$ for all $i$.  This is because our eventual solution will satisfy these bounds anyways, and omitting these bounds will drastically simplify our calculations.
We now rewrite \eqref{gradarg} as 
\begin{align}
 f \left ( \frac{\tilde{y}_i}{z_i} \right )  - \frac{\tilde{y}_i}{z_i} f' \left ( \frac{\tilde{y}_i}{z_i} \right ) + \lambda - \mu_1 & = 0  & i \in \mathcal{W}\label{wanted_grad}\\
 f \left ( \frac{\tilde{y}_i}{z_i} \right ) - \frac{\tilde{y}_i}{z_i} f' \left ( \frac{\tilde{y}_i}{z_i} \right ) + \lambda + \mu_2 & = 0& i \in \mathcal{U}\label{unwanted_grad}\\
 f \left ( \frac{\tilde{y}_i}{z_i} \right ) - \frac{\tilde{y}_i}{z_i} f' \left ( \frac{\tilde{y}_i}{z_i} \right ) + \lambda & = 0& i \in \mathcal{N}\label{neutral_grad}
 \end{align}

We now propose a solution can be found where that the ratios $\frac{\tilde{y}_i}{z_i}$ are constant in each of the sets $\mathcal{W}$, $\mathcal{U}$, $\mathcal{N}$.  That is 
\begin{align*}
z_i &= C_\mathcal{W}\tilde{y}_i & i \in \mathcal{W}\\
z_i &= C_\mathcal{U}\tilde{y}_i & i \in \mathcal{U}\\
z_i &= C_\mathcal{N}\tilde{y}_i & i \in \mathcal{N}.
\end{align*}

In that case we can satisfy conditions \eqref{wanted_grad}, \eqref{unwanted_grad} and \eqref{neutral_grad} (originally \eqref{gradarg}) by setting
\begin{align*}
\lambda &= C_\mathcal{N}^{-1}f'(C_\mathcal{N}^{-1})-f(C_\mathcal{N}^{-1})\\
\mu_1 &= \lambda + f(C_\mathcal{W}^{-1}) - C_\mathcal{W}^{-1} f'(C_\mathcal{W}^{-1})\\
\mu_2 &= -\lambda - f(C_\mathcal{U}^{-1})+C_\mathcal{U}^{-1} f'(C_\mathcal{U}^{-1}).
\end{align*}

We can now reformulate \eqref{d_feasibility} so that it is easier to analyze.  We will first define $h(x) = xf'(x)-f(x)$.  Note that because $f(x)$ is convex $h'(x) = xf''(x) \geq 0$ for all $x \geq 0$ and $h(x)$ is increasing.  We can then rewrite our formulas for $\lambda$, $\mu_1$ and $\mu_2$.
\begin{align*}
\lambda &= h(C_\mathcal{N}^{-1})\\
\mu_1 &= h(C_\mathcal{N}^{-1}) -h(C_\mathcal{W}^{-1})\\
\mu_2 &= h(C_\mathcal{U}^{-1}) - h(C_\mathcal{N}^{-1})
\end{align*}

Then $\mu_1 \geq 0$ becomes
\begin{align*}
h(C_\mathcal{N}^{-1}) &\geq h(C_\mathcal{W}^{-1})\\
C_\mathcal{N}^{-1} &\geq C_\mathcal{W}^{-1}\\
C_\mathcal{N} &\leq C_\mathcal{W},
\end{align*}
and $\mu_2 \geq 0$ similarly becomes $C_\mathcal{N} \geq C_\mathcal{U}$.  This means \eqref{d_feasibility} is equivalent to
\begin{equation}\label{d_simp_feasibility}
     C_\mathcal{U} \leq C_\mathcal{N} \leq C_\mathcal{W}
\end{equation}

\begin{figure}[h]
\begin{center}
\includegraphics[width = 0.5\textwidth]{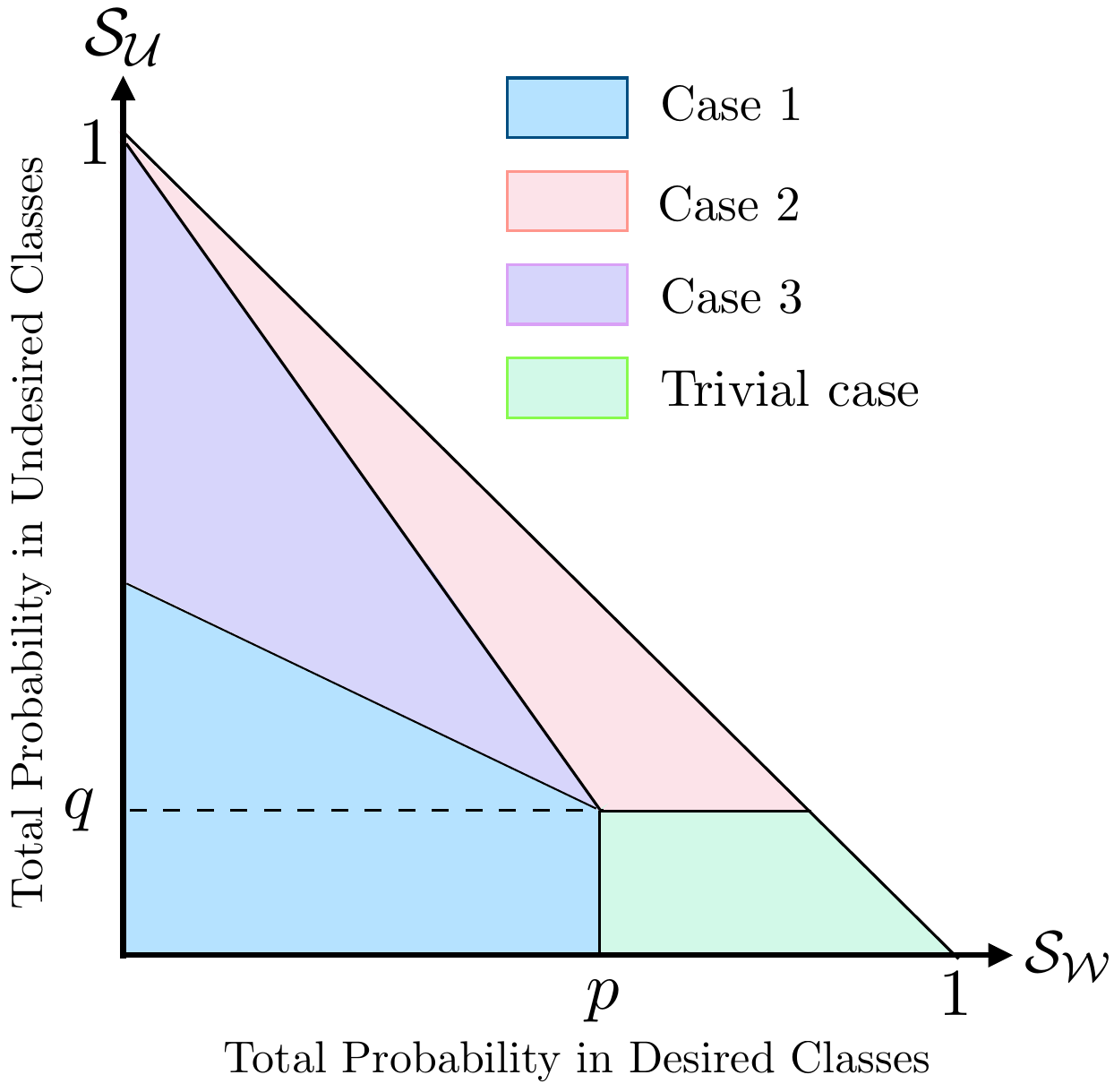}
\end{center}
\caption{The three cases visualized in probability space.}\label{img:cases}
\end{figure}

We must now find values of $C_\mathcal{W}$, $C_\mathcal{U}$ and $C_\mathcal{N}$ that satisfy \eqref{prob_cond} through \eqref{comp_unwanted}.  We will consider 3 cases illustrated in Figure \ref{img:cases}.

\textbf{Case: 1} Suppose $\mathcal{S}_\mathcal{W} < p$ and $\mathcal{S}_\mathcal{U}\leq (1-\mathcal{S}_\mathcal{W})\left ( \frac{q}{1-p} \right)$.

Let $C_\mathcal{W} = \frac{p}{\mathcal{S}_\mathcal{W}}$ and $C_\mathcal{U} = C_\mathcal{N} = \frac{1-p}{\mathcal{S}_\mathcal{U} + \mathcal{S}_\mathcal{N}}$.  This implies $\mu_2 = 0$ which satisfies \eqref{comp_unwanted} and half of \eqref{d_feasibility}.  This also implies $\sum_{i=\in \mathcal{W}} z_i = p$ satisfying \eqref{wanted_cond} and \eqref{comp_wanted}.  We will use the fact $\mathcal{S}_\mathcal{U} + \mathcal{S}_\mathcal{N} = 1-\mathcal{S}_\mathcal{W}$ in our proof of condition \eqref{unwanted_cond}.
\begin{align*}
\sum_{i\in \mathcal{S}_\mathcal{U}} z_i &= \sum_{i\in \mathcal{S}_\mathcal{U}} C_\mathcal{U} \tilde{y}_i= \frac{1-p}{\mathcal{S}_\mathcal{U} + \mathcal{S}_\mathcal{N}}\mathcal{S}_\mathcal{U}\\
&\leq \frac{1-p}{\mathcal{S}_\mathcal{U} + \mathcal{S}_\mathcal{N}} (1-\mathcal{S}_\mathcal{W})\left ( \frac{q}{1-p} \right) =q
\end{align*}

This proves \eqref{unwanted_cond} is satisfied.

Because $\mathcal{S}_\mathcal{W} < p$ we have 
\begin{align*}
    C_\mathcal{W}&=\frac{p}{\mathcal{S}_\mathcal{W}}> 1> \frac{1-p}{1-\mathcal{S}_\mathcal{W}}= \frac{1-p}{\mathcal{S}_\mathcal{U} + \mathcal{S}_\mathcal{N}}= C_\mathcal{N}
\end{align*}
This implies $\mu_1>0$ and satisfies the other half of \eqref{d_feasibility}.

We have now shown all the KKT conditions are satisfied and we have found a minimizer.  We now plug these values into \eqref{prob_dist} to find a closed form for the distance.
\begin{align*}
d_{\mathcal{Y}}(\tilde{\mathbf{y}},T) &=  \min_{\mathbf{z}\in T} \sum_{i=1}^k{z}_i f\left  (\frac{\tilde{y}_i}{z_i}\right )\\
&= \sum_{i\in \mathcal{W}}  \frac{p\tilde{y}_i}{\mathcal{S}_\mathcal{W}} f\left  (\frac{\mathcal{S}_\mathcal{W}}{p}\right ) + \sum_{i \notin \mathcal{W} }  \frac{(1-p)\tilde{y}_i}{\mathcal{S}_\mathcal{U}+ \mathcal{S}_\mathcal{N}} f\left  (\frac{\mathcal{S}_\mathcal{U}+ \mathcal{S}_\mathcal{N}}{1-p}\right )\\
&= pf\left( \frac{\mathcal{S}_\mathcal{W}}{p} \right ) + (1-p)f\left( \frac{\mathcal{S}_\mathcal{U}+\mathcal{S}_\mathcal{N}}{1-p} \right ).\\
\end{align*}

\textbf{Case: 2} Suppose $\mathcal{S}_\mathcal{U} > q$ and $\mathcal{S}_\mathcal{W} \geq (1-\mathcal{S}_\mathcal{U})\left ( \frac{p}{1-q} \right)$.

Let $C_\mathcal{U} = \frac{q}{\mathcal{S}_\mathcal{U}}$ and $C_\mathcal{W} = C_\mathcal{N} = \frac{1-q}{\mathcal{S}_\mathcal{W} + \mathcal{S}_\mathcal{N}}$.  This implies $\mu_1 = 0$ which satisfies \eqref{comp_wanted} and half of \eqref{d_feasibility}.  We also have $\sum_{i=\in \mathcal{U}} z_i = q$ satisfying \eqref{unwanted_cond} and \eqref{comp_unwanted}.   We now prove condition \eqref{wanted_cond} is satisfied.
\begin{align*}
\sum_{i\in \mathcal{S}_\mathcal{W}} z_i &= \sum_{i\in \mathcal{S}_\mathcal{W}} C_\mathcal{W} \tilde{y}_i= \frac{1-q}{\mathcal{S}_\mathcal{W} + \mathcal{S}_\mathcal{N}}\mathcal{S}_\mathcal{W}\\
&\geq \frac{1-q}{\mathcal{S}_\mathcal{W} + \mathcal{S}_\mathcal{N}} (1-\mathcal{S}_\mathcal{U})\left ( \frac{p}{1-q} \right)&=p
\end{align*}

Finally we prove $C_\mathcal{N} \geq C_\mathcal{U}$ implying  $\mu_2 \geq 0$ which satisfies the other half of \eqref{d_feasibility}
\begin{align*}
    C_\mathcal{U}&=\frac{q}{\mathcal{S}_\mathcal{U}} < 1 < \frac{1-q}{1-\mathcal{S}_\mathcal{U}}
    = \frac{1-q}{\mathcal{S}_\mathcal{W} + \mathcal{S}_\mathcal{N}}= C_\mathcal{N}
\end{align*}

Now that we have proven that this is a minimizer we will again plug solution into \eqref{prob_dist} to find the distance value.
\begin{align*}
d_{\mathcal{Y}}(\tilde{\mathbf{y}},T) &=  \min_{\mathbf{z}\in T} \sum_{i=1}^k{z}_i f\left  (\frac{\tilde{y}_i}{z_i}\right )\\
&= \sum_{i\in \mathcal{U}}  \frac{q\tilde{y}_i}{\mathcal{S}_\mathcal{U}} f\left  (\frac{\mathcal{S}_\mathcal{U}}{q}\right ) + \sum_{i \notin \mathcal{U} }  \frac{(1-q)\tilde{y}_i}{\mathcal{S}_\mathcal{W}+ \mathcal{S}_\mathcal{N}} f\left  (\frac{\mathcal{S}_\mathcal{W}+ \mathcal{S}_\mathcal{N}}{1-p}\right )\\
&= qf\left( \frac{\mathcal{S}_\mathcal{U}}{q} \right ) + (1-q)f\left( \frac{\mathcal{S}_\mathcal{W}+\mathcal{S}_\mathcal{N}}{1-q} \right ).\\
\end{align*}

\textbf{Case: 3} Suppose $\mathcal{S}_\mathcal{U}>(1-\mathcal{S}_\mathcal{W})\left ( \frac{q}{1-p} \right)$ and $\mathcal{S}_\mathcal{W}<(1-\mathcal{S}_\mathcal{U})\left ( \frac{p}{1-q} \right)$.

Let $C_\mathcal{W} = \frac{p}{\mathcal{S}_\mathcal{W}}$, $C_\mathcal{U} = \frac{q}{\mathcal{S}_\mathcal{U}}$ and $C_\mathcal{N} = \frac{1-p-q}{\mathcal{S}_\mathcal{N}}$ in which case $\sum_{i=\in \mathcal{W}} z_i = p$ (satisfying \eqref{wanted_cond} and \eqref{comp_wanted}), $\sum_{i=\in \mathcal{U}} z_i = q$ (satisfying \eqref{unwanted_cond} and \eqref{comp_unwanted}).  The choice of $C_\mathcal{N}$ ensures that \eqref{prob_cond} is satisfied:
\begin{align*}
    \sum_{i=1}^M z_i& = \sum_{i\in C_\mathcal{W}} z_i +\sum_{i\in C_\mathcal{U}} z_i + \sum_{i\in C_\mathcal{N}} z_i\\
    & = \sum_{i\in C_\mathcal{W}} C_\mathcal{W}\tilde{y}_i +\sum_{i\in C_\mathcal{U}} C_\mathcal{U}\tilde{y}_i  + \sum_{i\in C_\mathcal{N}} C_\mathcal{N}\tilde{y}_i \\
    & = C_\mathcal{W}\mathcal{S}_\mathcal{W} + C_\mathcal{U}\mathcal{S}_\mathcal{U} + C_\mathcal{N}\mathcal{S}_\mathcal{N}\\
    &= 1
\end{align*}

To show that \eqref{d_feasibility} is satisfied.  We note $\mathcal{S}_\mathcal{U}>(1-\mathcal{S}_\mathcal{W})\left ( \frac{q}{1-p} \right)$ implies $C_\mathcal{N} > C_\mathcal{U}$ and $\mathcal{S}_\mathcal{W}<(1-\mathcal{S}_\mathcal{U})\left ( \frac{p}{1-q} \right)p$ implies $C_\mathcal{N} < C_\mathcal{W}$.  this proves \eqref{d_simp_feasibility} which is equivalent to \eqref{d_feasibility}
Plugging these minimizing values of $\mathbf{z}$ into \eqref{prob_dist} yields
\begin{align*}
d_{\mathcal{Y}}(\tilde{\mathbf{y}},T) &=  \min_{\mathbf{z}\in T} \sum_{i=1}^k{z}_i f\left  (\frac{\tilde{y}_i}{z_i}\right )\\
&= \sum_{i\in \mathcal{W}}  \frac{p\tilde{y}_i}{\mathcal{S}_\mathcal{W}} f\left  (\frac{\mathcal{S}_\mathcal{W}}{p}\right ) + \sum_{i\in \mathcal{U}}  \frac{q\tilde{y}_i}{\mathcal{S}_\mathcal{U}} f\left  (\frac{\mathcal{S}_\mathcal{U}}{q}\right ) + \sum_{i\in \mathcal{W}}  \frac{(1-p-q)\tilde{y}_i}{\mathcal{S}_\mathcal{N}} f\left  (\frac{\mathcal{S}_\mathcal{N}}{1-p-q}\right )\\
&= pf\left( \frac{\mathcal{S}_\mathcal{W}}{p} \right ) + qf\left( \frac{\mathcal{S}_\mathcal{U}}{q} \right )+(1-p-q)f\left( \frac{\mathcal{S}_\mathcal{N}}{1-p-q} \right ).
\end{align*}

This proves the closed form in equation (\ref{closed_form}) and we may now proceed to show that this function is continuous.  To prove continuity we need only show continuity the piece-wise boundaries which we will evaluate one at a time.

\textbf{Boundary 1:} $\mathcal{S}_\mathcal{W} = p$.  The two functions that share this boundary are $0$ and $pf\left( \frac{\mathcal{S}_\mathcal{W}}{p} \right ) + (1-p)f\left( \frac{1-\mathcal{S}_\mathcal{W}}{1-p} \right )$.  Plugging the boundary into the latter function yields
\begin{align*}
    pf\left( \frac{\mathcal{S}_\mathcal{W}}{p} \right ) + (1-p)f\left( \frac{1-\mathcal{S}_\mathcal{W}}{1-p} \right ) &= pf\left( \frac{p}{p} \right ) + (1-p)f\left( \frac{1-p}{1-p} \right )\\
    &= 0.
\end{align*}

The two functions are equal on the boundary and the boundary is continuous.

\textbf{Boundary 2:} $\mathcal{S}_\mathcal{U} = q$.  The two functions that share this boundary are $0$ and $qf\left( \frac{\mathcal{S}_\mathcal{U}}{q} \right ) + (1-q)f\left( \frac{1-\mathcal{S}_\mathcal{U}}{1-q} \right )$.  Plugging the boundary into the latter function yields
\begin{align*}
    qf\left( \frac{\mathcal{S}_\mathcal{U}}{q} \right ) + (1-q)f\left( \frac{1-\mathcal{S}_\mathcal{U}}{1-q} \right ) &= qf\left( \frac{q}{q} \right ) + (1-q)f\left( \frac{1-q}{1-q} \right )\\
    &= 0.
\end{align*}

The two functions are equal on the boundary and the boundary is continuous.

\textbf{Boundary 3:} $\mathcal{S}_\mathcal{U} = (1-\mathcal{S}_\mathcal{W})\left ( \frac{q}{1-p} \right)$.  The two functions that share this boundary are $pf\left( \frac{\mathcal{S}_\mathcal{W}}{p} \right ) + (1-p)f\left( \frac{1-\mathcal{S}_\mathcal{W}}{1-p} \right )$ and $pf\left( \frac{\mathcal{S}_\mathcal{W}}{p} \right ) + qf\left( \frac{\mathcal{S}_\mathcal{U}}{q} \right )+(1-p-q)f\left( \frac{\mathcal{S}_\mathcal{N}}{1-p-q} \right )$.  Plugging the boundary into the latter function yields
\begin{align*}
    pf\left( \frac{\mathcal{S}_\mathcal{W}}{p} \right ) + qf\left( \frac{\mathcal{S}_\mathcal{U}}{q} \right )+(1-p-q)f\left( \frac{1-\mathcal{S}_\mathcal{W}-\mathcal{S}_\mathcal{U}}{1-p-q} \right ) =& pf\left( \frac{\mathcal{S}_\mathcal{W}}{p} \right ) + (1-p)f\left( \frac{1-\mathcal{S}_\mathcal{W}}{1-p} \right ).
\end{align*}

The two functions are equal on the boundary and the boundary is continuous.

\textbf{Boundary 4:} $\mathcal{S}_\mathcal{W} = (1-\mathcal{S}_\mathcal{U})\left ( \frac{p}{1-q} \right)$.  The two functions that share this boundary are $qf\left( \frac{\mathcal{S}_\mathcal{U}}{q} \right ) + (1-q)f\left( \frac{1-\mathcal{S}_\mathcal{U}}{1-q} \right )$ and $pf\left( \frac{\mathcal{S}_\mathcal{W}}{p} \right ) + qf\left( \frac{\mathcal{S}_\mathcal{U}}{q} \right )+(1-p-q)f\left( \frac{\mathcal{S}_\mathcal{N}}{1-p-q} \right )$.  Plugging the boundary into the latter function yields
\begin{align*}
    qf\left( \frac{\mathcal{S}_\mathcal{U}}{q} \right ) + pf\left( \frac{\mathcal{S}_\mathcal{W}}{p} \right ) + (1-p-q)f\left( \frac{1-\mathcal{S}_\mathcal{U}-\mathcal{S}_\mathcal{W}}{1-p-q} \right ) =& qf\left( \frac{\mathcal{S}_\mathcal{U}}{q} \right ) + (1-q)f\left( \frac{1-\mathcal{S}_\mathcal{U}}{1-q} \right ).
\end{align*}

The two functions are equal on the boundary and the boundary is continuous.  We have now shown continuity on all boundaries and the function is continuous.  Now to show that the derivative of the function is continuous we need only show the all partial derivatives exist and agree on the boundaries.  We use the closed form equation \eqref{closed_form} found in the body of the paper (which is equivalent to the one found in the beginning of the proof) but suppresses $\mathcal{S}_\mathcal{N}$.  This makes it easier to differentiate with respect to $\tilde{y}_i$, $i\in \mathcal{W} \cup \mathcal{U}$. 
\begin{equation*}
d_{\mathcal{Y}}(\tilde{\mathbf{y}},T)=
    \begin{cases}
        0 & \text{if } \mathcal{S}_\mathcal{W}\geq p \text{ and } \mathcal{S}_\mathcal{U} \leq q\\
        
        pf\left( \frac{\mathcal{S}_\mathcal{W}}{p} \right ) + (1-p)f\left( \frac{1-\mathcal{S}_\mathcal{W}}{1-p} \right )& \text{if } \mathcal{S}_\mathcal{W} < p \text{ and } \mathcal{S}_\mathcal{U}\leq (1-\mathcal{S}_\mathcal{W})\left ( \frac{q}{1-p} \right)\\
        
        qf\left( \frac{\mathcal{S}_\mathcal{U}}{q} \right )+(1-q)f\left( \frac{1-\mathcal{S}_\mathcal{U}}{1-q} \right )& \text{if } \mathcal{S}_\mathcal{U} > q \text{ and } \mathcal{S}_\mathcal{W} \geq (1-\mathcal{S}_\mathcal{U})\left ( \frac{p}{1-q} \right)\\

        pf\left( \frac{\mathcal{S}_\mathcal{W}}{p} \right ) + qf\left( \frac{\mathcal{S}_\mathcal{U}}{q} \right )+(1-p-q)f\left( \frac{1- \mathcal{S}_\mathcal{W}-\mathcal{S}_\mathcal{U}}{1-p-q} \right )& \text{if } \mathcal{S}_\mathcal{U}>(1-\mathcal{S}_\mathcal{W})\left ( \frac{q}{1-p} \right) \\
        &\text{ and } \mathcal{S}_\mathcal{W}<(1-\mathcal{S}_\mathcal{U})\left ( \frac{p}{1-q} \right)
    \end{cases}
\end{equation*}

We now take the derivative with respect to a desirable class ($i\in \mathcal{W}$).

\begin{equation*}
\frac{\partial}{\partial \tilde{y}_{i\in \mathcal{W}}}d_{\mathcal{Y}}(\tilde{\mathbf{y}},T)=
    \begin{cases}
        0 & \text{if } \mathcal{S}_\mathcal{W}> p \text{ and } \mathcal{S}_\mathcal{U} < q\\
        
        f'\left( \frac{\mathcal{S}_\mathcal{W}}{p} \right ) - f'\left( \frac{1-\mathcal{S}_\mathcal{W}}{1-p} \right )& \text{if } \mathcal{S}_\mathcal{W} < p \text{ and } \mathcal{S}_\mathcal{U}< (1-\mathcal{S}_\mathcal{W})\left ( \frac{q}{1-p} \right)\\
        
        0& \text{if } \mathcal{S}_\mathcal{U} > q \text{ and } \mathcal{S}_\mathcal{W} > (1-\mathcal{S}_\mathcal{U})\left ( \frac{p}{1-q} \right)\\

        f'\left( \frac{\mathcal{S}_\mathcal{W}}{p} \right )-f'\left( \frac{1- \mathcal{S}_\mathcal{W}-\mathcal{S}_\mathcal{U}}{1-p-q} \right )& \text{if } \mathcal{S}_\mathcal{U}>(1-\mathcal{S}_\mathcal{W})\left ( \frac{q}{1-p} \right) \\
        &\text{ and } \mathcal{S}_\mathcal{W}<(1-\mathcal{S}_\mathcal{U})\left ( \frac{p}{1-q} \right)
    \end{cases}
\end{equation*}
Now we need only ensure all pieces agree on the boundaries to show that the derivative exists and is continuous.

\textbf{Boundary 1:} $\mathcal{S}_\mathcal{W} = p$.  The two functions that share this boundary are $0$ and $ f'\left( \frac{\mathcal{S}_\mathcal{W}}{p} \right ) - f'\left( \frac{1-\mathcal{S}_\mathcal{W}}{1-p} \right )$.  Plugging the boundary into the latter function yields
\begin{align*}
    f'\left( \frac{\mathcal{S}_\mathcal{W}}{p} \right ) - f'\left( \frac{1-\mathcal{S}_\mathcal{W}}{1-p} \right ) &= f'\left( \frac{p}{p} \right ) - f'\left( \frac{1-p}{1-p} \right )\\
    &= f'(1) + f(1')\\
    &= 0.
\end{align*}
Then setting the derivative at the boundary to $0$ makes the derivative on this boundary continuous.

\textbf{Boundary 2:} $\mathcal{S}_\mathcal{U} = q$.  The two functions that share this boundary are both $0$, and setting the derivative at the boundary to $0$ makes the derivative on this boundary continuous.

\textbf{Boundary 3:} $\mathcal{S}_\mathcal{U} = (1-\mathcal{S}_\mathcal{W})\left ( \frac{q}{1-p} \right)$.  The two functions that share this boundary are $ f'\left( \frac{\mathcal{S}_\mathcal{W}}{p} \right ) - f'\left( \frac{1-\mathcal{S}_\mathcal{W}}{1-p} \right )$ and $f'\left( \frac{\mathcal{S}_\mathcal{W}}{p} \right )-f'\left( \frac{1- \mathcal{S}_\mathcal{W}-\mathcal{S}_\mathcal{U}}{1-p-q} \right )$.  Plugging the boundary into the latter function yields
\begin{align*}
  f'\left( \frac{\mathcal{S}_\mathcal{W}}{p} \right )-f'\left( \frac{1- \mathcal{S}_\mathcal{W}-\mathcal{S}_\mathcal{U}}{1-p-q} \right ) =& f'\left( \frac{\mathcal{S}_\mathcal{W}}{p} \right )-f'\left( \frac{1- \mathcal{S}_\mathcal{W}}{1-p} \right )
\end{align*}

Then setting the derivative at the boundary to $f'\left( \frac{\mathcal{S}_\mathcal{W}}{p} \right )-f'\left( \frac{1- \mathcal{S}_\mathcal{W}}{1-p} \right )$ makes the derivative on this boundary continuous.

\textbf{Boundary 4:} $\mathcal{S}_\mathcal{W} = (1-\mathcal{S}_\mathcal{U})\left ( \frac{p}{1-q} \right)$.  The two functions that share this boundary are $0$ and $ f'\left( \frac{\mathcal{S}_\mathcal{W}}{p} \right )-f'\left( \frac{1- \mathcal{S}_\mathcal{W}-\mathcal{S}_\mathcal{U}}{1-p-q} \right )$.  We rewrite the boundary as $\mathcal{S}_\mathcal{U} =\frac{1-q}{p}\mathcal{S}_\mathcal{W} +1$ and plug it into the latter function.
\begin{align*}
     f'\left( \frac{\mathcal{S}_\mathcal{W}}{p} \right )-f'\left( \frac{1- \mathcal{S}_\mathcal{W}-\mathcal{S}_\mathcal{U}}{1-p-q} \right ) =& f'\left( \frac{\mathcal{S}_\mathcal{W}}{p} \right )-f'\left( \frac{1- \mathcal{S}_\mathcal{W}-\left ( \frac{1-q}{p}\mathcal{S}_\mathcal{W} +1 \right )}{1-p-q} \right )\\
     =& 0
\end{align*}
Then setting the derivative at the boundary to $0$ makes the derivative on this boundary continuous.

This yields the continuous partial derivative
\begin{equation}\label{desirable_derivative}
\frac{\partial}{\partial \tilde{y}_{i\in \mathcal{W}}}d_{\mathcal{Y}}(\tilde{\mathbf{y}},T)=
    \begin{cases}
        0 & \text{if } \mathcal{S}_\mathcal{W}\geq p \text{ and } \mathcal{S}_\mathcal{U} \leq q\\
        
        f'\left( \frac{\mathcal{S}_\mathcal{W}}{p} \right ) - f'\left( \frac{1-\mathcal{S}_\mathcal{W}}{1-p} \right )& \text{if } \mathcal{S}_\mathcal{W} < p \text{ and } \mathcal{S}_\mathcal{U}\leq (1-\mathcal{S}_\mathcal{W})\left ( \frac{q}{1-p} \right)\\
        
        0& \text{if } \mathcal{S}_\mathcal{U} > q \text{ and } \mathcal{S}_\mathcal{W} \geq (1-\mathcal{S}_\mathcal{U})\left ( \frac{p}{1-q} \right)\\

        f'\left( \frac{\mathcal{S}_\mathcal{W}}{p} \right )-f'\left( \frac{1- \mathcal{S}_\mathcal{W}-\mathcal{S}_\mathcal{U}}{1-p-q} \right )& \text{if } \mathcal{S}_\mathcal{U}>(1-\mathcal{S}_\mathcal{W})\left ( \frac{q}{1-p} \right) \\
        &\text{ and } \mathcal{S}_\mathcal{W}<(1-\mathcal{S}_\mathcal{U})\left ( \frac{p}{1-q} \right)
    \end{cases}.
\end{equation} 

We now take the derivative with respect to a undesirable class ($i\in \mathcal{U}$).

\begin{equation*}
\frac{\partial}{\partial \tilde{y}_{i\in \mathcal{U}}}d_{\mathcal{Y}}(\tilde{\mathbf{y}},T)=
    \begin{cases}
        0 & \text{if } \mathcal{S}_\mathcal{W}> p \text{ and } \mathcal{S}_\mathcal{U} < q\\
        
        0& \text{if } \mathcal{S}_\mathcal{W} < p \text{ and } \mathcal{S}_\mathcal{U}< (1-\mathcal{S}_\mathcal{W})\left ( \frac{q}{1-p} \right)\\
        
        f'\left ( \frac{\mathcal{S}_\mathcal{U}}{q} \right ) - f'\left ( \frac{1-\mathcal{S}_\mathcal{U}}{1-q} \right ) & \text{if } \mathcal{S}_\mathcal{U} > q \text{ and } \mathcal{S}_\mathcal{W} > (1-\mathcal{S}_\mathcal{U})\left ( \frac{p}{1-q} \right)\\

        f'\left( \frac{\mathcal{S}_\mathcal{U}}{q} \right )-f'\left( \frac{1- \mathcal{S}_\mathcal{W}-\mathcal{S}_\mathcal{U}}{1-p-q} \right )& \text{if } \mathcal{S}_\mathcal{U}>(1-\mathcal{S}_\mathcal{W})\left ( \frac{q}{1-p} \right) \\
        &\text{ and } \mathcal{S}_\mathcal{W}<(1-\mathcal{S}_\mathcal{U})\left ( \frac{p}{1-q} \right)
    \end{cases}
\end{equation*}
Now we need only ensure that there is agreement on the boundaries.

\textbf{Boundary 1:} $\mathcal{S}_\mathcal{W} = p$.  The two functions that share this boundary are both $0$, and setting the derivative at the boundary to $0$ makes the derivative on this boundary continuous.

\textbf{Boundary 2:} $\mathcal{S}_\mathcal{U} = q$.  The two functions that share this boundary are both $0$ and $f'\left ( \frac{\mathcal{S}_\mathcal{U}}{q} \right ) - f'\left ( \frac{1-\mathcal{S}_\mathcal{U}}{1-q} \right )$.  Plugging the boundary into the latter function yields
\begin{align*}
    f'\left( \frac{\mathcal{S}_\mathcal{U}}{q} \right ) - f'\left( \frac{1-\mathcal{S}_\mathcal{U}}{1-q} \right ) &= f'\left( \frac{q}{q} \right ) - f'\left( \frac{1-q}{1-q} \right )\\
    &= 0.
\end{align*}
Then setting the derivative at the boundary to $0$ makes the derivative on this boundary continuous.

\textbf{Boundary 3:} $\mathcal{S}_\mathcal{U} = (1-\mathcal{S}_\mathcal{W})\left ( \frac{q}{1-p} \right)$.  The two functions that share this boundary are $0$ and $f'\left( \frac{\mathcal{S}_\mathcal{U}}{q} \right )-f'\left( \frac{1- \mathcal{S}_\mathcal{W}-\mathcal{S}_\mathcal{U}}{1-p-q} \right )$.  We rewrite the boundary as $\mathcal{S}_\mathcal{W} =1 - \frac{1-p}{q}\mathcal{S}_\mathcal{U}$ and  plug it into the latter function.
\begin{align*}
  f'\left( \frac{\mathcal{S}_\mathcal{U}}{q} \right )-f'\left( \frac{1- \mathcal{S}_\mathcal{W}-\mathcal{S}_\mathcal{U}}{1-p-q} \right ) =& f'\left( \frac{\mathcal{S}_\mathcal{U}}{q} \right )-f'\left( \frac{1- \mathcal{S}_\mathcal{U}-\left (1- \frac{1-p}{q}\mathcal{S}_\mathcal{U} \right )}{1-p-q} \right )\\
  =& 0
\end{align*}

Then setting the derivative at the boundary to $0$ makes the derivative on this boundary continuous.

\textbf{Boundary 4:} $\mathcal{S}_\mathcal{W} = (1-\mathcal{S}_\mathcal{U})\left ( \frac{p}{1-q} \right)$.  The two functions that share this boundary are $f'\left ( \frac{\mathcal{S}_\mathcal{U}}{q} \right ) - f'\left ( \frac{1-\mathcal{S}_\mathcal{U}}{1-q} \right )$ and  $f'\left( \frac{\mathcal{S}_\mathcal{U}}{q} \right )-f'\left( \frac{1- \mathcal{S}_\mathcal{W}-\mathcal{S}_\mathcal{U}}{1-p-q} \right )$.  Plugging the boundary into the latter function yields
\begin{align*}
     f'\left( \frac{\mathcal{S}_\mathcal{U}}{q} \right )-f'\left( \frac{1- \mathcal{S}_\mathcal{W}-\mathcal{S}_\mathcal{U}}{1-p-q} \right ) =& f'\left( \frac{\mathcal{S}_\mathcal{U}}{q} \right )-f'\left( \frac{1- \mathcal{S}_\mathcal{U}-(1-\mathcal{S}_\mathcal{U})\left ( \frac{p}{1-q} \right)}{1-p-q} \right )\\
    =&f'\left ( \frac{\mathcal{S}_\mathcal{U}}{q} \right ) -f'\left ( \frac{1-\mathcal{S}_\mathcal{U}}{1-q} \right )
\end{align*}
Then setting the derivative at the boundary to $f'\left ( \frac{\mathcal{S}_\mathcal{U}}{q} \right )-f'\left ( \frac{1-\mathcal{S}_\mathcal{U}}{1-q} \right )$ makes the derivative on this boundary continuous.

This yields the continuous partial derivative
\begin{equation}\label{undesirable_derivative}
\frac{\partial}{\partial \tilde{y}_{i\in \mathcal{U}}}d_{\mathcal{Y}}(\tilde{\mathbf{y}},T)=
    \begin{cases}
        0 & \text{if } \mathcal{S}_\mathcal{W}\geq p \text{ and } \mathcal{S}_\mathcal{U} \leq q\\
        
        0& \text{if } \mathcal{S}_\mathcal{W} < p \text{ and } \mathcal{S}_\mathcal{U}\leq (1-\mathcal{S}_\mathcal{W})\left ( \frac{q}{1-p} \right)\\
        
        f'\left ( \frac{\mathcal{S}_\mathcal{U}}{q} \right ) - f'\left ( \frac{1-\mathcal{S}_\mathcal{U}}{1-q} \right ) & \text{if } \mathcal{S}_\mathcal{U} > q \text{ and } \mathcal{S}_\mathcal{W} \geq (1-\mathcal{S}_\mathcal{U})\left ( \frac{p}{1-q} \right)\\

        f'\left( \frac{\mathcal{S}_\mathcal{U}}{q} \right )-f'\left( \frac{1- \mathcal{S}_\mathcal{W}-\mathcal{S}_\mathcal{U}}{1-p-q} \right )& \text{if } \mathcal{S}_\mathcal{U}>(1-\mathcal{S}_\mathcal{W})\left ( \frac{q}{1-p} \right) \\
        &\text{ and } \mathcal{S}_\mathcal{W}<(1-\mathcal{S}_\mathcal{U})\left ( \frac{p}{1-q} \right)
    \end{cases}.
\end{equation}

\textbf{Additional Analysis on $d_\mathcal{Y}$}
The following lemma shows that $d_\mathcal{Y}$ exhibits desirable behavior for any $f$-divergence if we restrict ourselves to the binary classification setting. 

\begin{lemma}\label{lemma_nondecreasing}
    In the binary classification setting, if $T = \{{\mathbf{z}}\in \mathcal{Y} | z_1 \geq p\},$ then $d_{\mathcal{Y}}(\tilde{\mathbf{y}},T)$ is decreasing (not necessarily strictly) in $\tilde{\mathbf{y}}_1$ for $D(\tilde{\mathbf{y}} || \mathbf{z})$ any $f$-divergence.
\end{lemma}


We now present the proof of Lemma \ref{lemma_nondecreasing}. Recall $d_{\mathcal{Y}}(\tilde{\mathbf{y}},T) = \min_{\mathbf{z}\in T} D_f(\tilde{\mathbf{y}}||\mathbf{z})$.  For binary probability distributions $\mathbf{a}$ and $\mathbf{b}$, the $f$-divergence has the simple form 
\begin{equation}
  D_f(\mathbf{b}||\mathbf{a}) = \mathbf{a}_1 f \left ( \frac{\mathbf{b}_1}{\mathbf{a}_1} \right ) + (1-\mathbf{a}_1) f \left ( \frac{1-\mathbf{b}_1}{1-\mathbf{a}_1} \right )
\end{equation}
for a convex function $f$ with $f(1) = 0$.
We show a relationship between this formula and a secant line.  To refer to the secant line of a function $g(x)$ from point $x = \alpha$ to $x = \beta$ evaluated at $\gamma$, we will use the notation $S_g(\alpha,\beta;\gamma)$.  When using this notation we will assume that $\alpha \leq \beta$.

We assume $\mathbf{a}_1 > \mathbf{b}_1$ and show that $D_f(\mathbf{b}||\mathbf{a})$ is equivalent to the secant line of $f(x)$ from $x = \frac{\mathbf{b}_1}{\mathbf{a}_1}$ to $\frac{1-\mathbf{b}_1}{1-\mathbf{a}_1}$ evaluated at $1$. (Note $\frac{\mathbf{b}_1}{\mathbf{a}_1}<1<\frac{1-\mathbf{b}_1}{1-\mathbf{a}_1}$.) We show this simply using the point slope form.
\begin{align*}
S_f\left (\frac{\mathbf{b}_1}{\mathbf{a}_1},\frac{1-\mathbf{b}_1}{1-\mathbf{a}_1};x \right ) &= \left ( x-\frac{1-\mathbf{b}_1}{1-\mathbf{a}_1} \right ) \frac{ f \left ( \frac{1-\mathbf{b}_1}{1-\mathbf{a}_1}\right) - f \left ( \frac{\mathbf{b}_1}{\mathbf{a}_1} \right ) }{\frac{1-\mathbf{b}_1}{1-\mathbf{a}_1}-\frac{\mathbf{b}_1}{\mathbf{a}_1}} + f \left ( \frac{1-\mathbf{b}_1}{1-\mathbf{a}_1} \right )\\
S_f\left (\frac{\mathbf{b}_1}{\mathbf{a}_1},\frac{1-\mathbf{b}_1}{1-\mathbf{a}_1};1 \right ) &= \left ( 1-\frac{1-\mathbf{b}_1}{1-\mathbf{a}_1} \right ) \frac{ f \left ( \frac{1-\mathbf{b}_1}{1-\mathbf{a}_1}\right) - f \left ( \frac{\mathbf{b}_1}{\mathbf{a}_1} \right ) }{\frac{1-\mathbf{b}_1}{1-\mathbf{a}_1}-\frac{\mathbf{b}_1}{\mathbf{a}_1}} + f \left ( \frac{1-\mathbf{b}_1}{1-\mathbf{a}_1} \right )\\
&= \mathbf{a}_1 f \left ( \frac{\mathbf{b}_1}{\mathbf{a}_1} \right ) + (1-\mathbf{a}_1) f \left ( \frac{1-\mathbf{b}_1}{1-\mathbf{a}_1} \right )\\
& = D_f(\mathbf{b}||\mathbf{a})
\end{align*}

Now that  $D_f(\mathbf{b}||\mathbf{a})$ is related to a secant line we prove a few facts about secant lines of convex functions.  If $g$ is convex, then $S_g(\alpha,\beta;\gamma)$ is decreasing in  $\alpha$ and increasing in $\beta$ whenever $\alpha<\gamma<\beta$.  Recall that if $g$ is convex, then by definition for any $v_1<v_2<v_3$, we have
\begin{equation}
\frac{g(v_2)-g(v_1)}{v_2-v_1} \leq \frac{g(v_3)-g(v_1)}{v_3-v_1} \leq  \frac{g(v_3)-g(v_2)}{v_3-v_2}.
\end{equation}
Then for any $\beta<\tilde{\beta}$ we have
\begin{align}
    S_g(\alpha,\beta;\gamma) &= (\gamma-\alpha)m+g(\alpha)\\
    S_g(\alpha,\tilde{\beta};\gamma) &= (\gamma-\alpha)\tilde{m}+g(\alpha)
\end{align}
for $\tilde{m}\geq m$.  It follows that for any $\gamma\geq\alpha$
\begin{equation}
    S_g(\alpha,\beta;x) \leq S_g(\alpha,\tilde{\beta};x),
\end{equation}
and $S_g(\alpha,\beta;x)$ is increasing in $\beta$.

A similar argument shows that $S_g(\alpha,\beta;x)$ is decreasing in $\alpha$ when $\gamma\leq\beta$.

We will use these facts to analyze $d_{\mathcal{Y}}(\tilde{\mathbf{y}},T) = \min_{\mathbf{z}\in T} D_f(\tilde{\mathbf{y}}||\mathbf{z})$.  The $f$-divergence between identical distributions is zero, so we have $d_{\mathcal{Y}}(\tilde{\mathbf{y}},T) = 0$ whenever $\tilde{\mathbf{y}}_1\geq p$.  When $\tilde{\mathbf{y}}_1 < p$ we have $\frac{\tilde{\mathbf{y}}_1}{\mathbf{z}_1}<1<\frac{1-\tilde{\mathbf{y}}_1}{1-\mathbf{z}_1}$ and
\begin{align*}
    d_{\mathcal{Y}}(\tilde{\mathbf{y}},T) &= \min_{\mathbf{z}\in T} D_f(\tilde{\mathbf{y}}||\mathbf{z})\\
    &= \min_{\mathbf{z}\in T}S_f\left (\frac{\tilde{\mathbf{y}}_1}{\mathbf{z}_1},\frac{1-\tilde{\mathbf{y}}_1}{1-\mathbf{z}_1};1\right ),
\end{align*}
which is decreasing in $\frac{\tilde{\mathbf{y}}_1}{\mathbf{z}_1}$ and increasing in $\frac{1-\tilde{\mathbf{y}}_1}{1-\mathbf{z}_1}$, so to achieve the minimum we use the smallest possible $\mathbf{z}_1$, i.e. $\mathbf{z}_1 = p$.  We may now simplify
\begin{equation*}
d_{\mathcal{Y}}(\tilde{\mathbf{y}},T) =
    \begin{cases}
        S_f\left (\frac{\tilde{\mathbf{y}}_1}{p},\frac{1-\tilde{\mathbf{y}}_1}{1-p};1\right ) & \text{if } \tilde{\mathbf{y}} < p \\
        0 & \text{if } \tilde{\mathbf{y}} \geq p
    \end{cases}.
\end{equation*}
Note that this is continuous at $\tilde{\mathbf{y}} =p$ because $S_f (1,1;1) = f(1) = 0$.  With this closed form solution for $d_{\mathcal{Y}}(\tilde{\mathbf{y}},T)$ we may finish the proof.

  We have already shown that $S_f\left (\frac{\tilde{\mathbf{y}}_1}{p},\frac{1-\tilde{\mathbf{y}}_1}{1-p};1\right )$ is decreasing in $\frac{\tilde{\mathbf{y}}_1}{p}$ and increasing in $\frac{1-\tilde{\mathbf{y}}_1}{1-p}$, so increasing $\tilde{\mathbf{y}}_1$ will decrease $S_f\left (\frac{\tilde{\mathbf{y}}_1}{p},\frac{1-\tilde{\mathbf{y}}_1}{1-p};1\right )$ and $d_{\mathcal{Y}}(\tilde{\mathbf{y}},T)$ is decreasing in $\tilde{\mathbf{y}}_1$.

We now present a corollary to Theorem \ref{theorem_closedform} that shows explicitly that $d_\mathcal{Y}$ decreases with added probability to the desirable classes and increases with added probability to the undesirable classes.

\begin{corollary}\label{general_properties}
    If $T$ is of form \eqref{gen_target_fam} and $f$ is twice differentiable, then $d_{\mathcal{Y}}(\tilde{\mathbf{y}},T)$ is decreasing in $\tilde{\mathbf{y}}_i$ if $i\in\mathcal{W}$ and is increasing if $i\in\mathcal{U}$.
\end{corollary}

To prove Corollary \ref{general_properties}, we need only show equation \eqref{closed_form} is decreasing in $\tilde{y}_i$ for $i\in \mathcal{W}$ and increasing in $\tilde{y}_i$ for $i\in \mathcal{U}$, we need only prove that the partial derivative \eqref{desirable_derivative} is non-positive and the partial derivative \eqref{undesirable_derivative} is non-negative.  We will rely heavily on the fact that$f'$ is increasing because $f$ is convex.

We start with \eqref{desirable_derivative}:

\begin{equation*}
\frac{\partial}{\partial \tilde{y}_{i\in \mathcal{W}}}d_{\mathcal{Y}}(\tilde{\mathbf{y}},T)=
    \begin{cases}
        0 & \text{if } \mathcal{S}_\mathcal{W}\geq p \text{ and } \mathcal{S}_\mathcal{U} \leq q\\
        
        f'\left( \frac{\mathcal{S}_\mathcal{W}}{p} \right ) - f'\left( \frac{1-\mathcal{S}_\mathcal{W}}{1-p} \right )& \text{if } \mathcal{S}_\mathcal{W} < p \text{ and } \mathcal{S}_\mathcal{U}\leq (1-\mathcal{S}_\mathcal{W})\left ( \frac{q}{1-p} \right)\\
        
        0& \text{if } \mathcal{S}_\mathcal{U} > q \text{ and } \mathcal{S}_\mathcal{W} \geq (1-\mathcal{S}_\mathcal{U})\left ( \frac{p}{1-q} \right)\\

        f'\left( \frac{\mathcal{S}_\mathcal{W}}{p} \right )-f'\left( \frac{1- \mathcal{S}_\mathcal{W}-\mathcal{S}_\mathcal{U}}{1-p-q} \right )& \text{if } \mathcal{S}_\mathcal{U}>(1-\mathcal{S}_\mathcal{W})\left ( \frac{q}{1-p} \right) \\
        &\text{ and } \mathcal{S}_\mathcal{W}<(1-\mathcal{S}_\mathcal{U})\left ( \frac{p}{1-q} \right)
    \end{cases}.
\end{equation*} 

Clearly the first and third cases are non-positive, so we proceed to the second case.

Because $\mathcal{S}_\mathcal{W} < p$, we have $\frac{\mathcal{S}_\mathcal{W}}{p}<1<\frac{1-\mathcal{S}_\mathcal{W}}{1-p}$ and 
\begin{align*}
    f'\left ( \frac{\mathcal{S}_\mathcal{W}}{p}  \right ) &<f'\left ( \frac{1-\mathcal{S}_\mathcal{W}}{1-p}  \right )\\
    f'\left ( \frac{\mathcal{S}_\mathcal{W}}{p}  \right )-f'\left ( \frac{1-\mathcal{S}_\mathcal{W}}{1-p}  \right )&<0.
\end{align*}

Next we prove the partial derivative is negative in the fourth case.  
\begin{align*}
    \mathcal{S}_\mathcal{W}&<(1-\mathcal{S}_\mathcal{U})\left ( \frac{p}{1-q} \right)\\
    \mathcal{S}_\mathcal{W}-q\mathcal{S}_\mathcal{W}&<p-p\mathcal{S}_\mathcal{U}\\
    \mathcal{S}_\mathcal{W}-q\mathcal{S}_\mathcal{W}-p\mathcal{S}_\mathcal{W}&<p-p\mathcal{S}_\mathcal{U}-p\mathcal{S}_\mathcal{W}\\
    \frac{\mathcal{S}_\mathcal{W}}{p}&<\frac{1-\mathcal{S}_\mathcal{U}-\mathcal{S}_\mathcal{W}}{1-p-q}\\
    f'\left (\frac{\mathcal{S}_\mathcal{W}}{p}\right )&<f'\left ( \frac{1-\mathcal{S}_\mathcal{U}-\mathcal{S}_\mathcal{W}}{1-p-q}\right )\\
    f'\left (\frac{\mathcal{S}_\mathcal{W}}{p}\right ) - f'\left ( \frac{1-\mathcal{S}_\mathcal{U}-\mathcal{S}_\mathcal{W}}{1-p-q}\right )&<0
\end{align*}
This shows that \eqref{desirable_derivative} is non-positive and \eqref{closed_form} is decreasing in $\tilde{y}_i$ for $i\in \mathcal{W}$.

We now consider \eqref{undesirable_derivative}:
\begin{equation*}
\frac{\partial}{\partial \tilde{y}_{i\in \mathcal{U}}}d_{\mathcal{Y}}(\tilde{\mathbf{y}},T)=
    \begin{cases}
        0 & \text{if } \mathcal{S}_\mathcal{W}\geq p \text{ and } \mathcal{S}_\mathcal{U} \leq q\\
        
        0& \text{if } \mathcal{S}_\mathcal{W} < p \text{ and } \mathcal{S}_\mathcal{U}\leq (1-\mathcal{S}_\mathcal{W})\left ( \frac{q}{1-p} \right)\\
        
        f'\left ( \frac{\mathcal{S}_\mathcal{U}}{q} \right ) - f'\left ( \frac{1-\mathcal{S}_\mathcal{U}}{1-q} \right ) & \text{if } \mathcal{S}_\mathcal{U} > q \text{ and } \mathcal{S}_\mathcal{W} \geq (1-\mathcal{S}_\mathcal{U})\left ( \frac{p}{1-q} \right)\\

        f'\left( \frac{\mathcal{S}_\mathcal{U}}{q} \right )-f'\left( \frac{1- \mathcal{S}_\mathcal{W}-\mathcal{S}_\mathcal{U}}{1-p-q} \right )& \text{if } \mathcal{S}_\mathcal{U}>(1-\mathcal{S}_\mathcal{W})\left ( \frac{q}{1-p} \right) \\
        &\text{ and } \mathcal{S}_\mathcal{W}<(1-\mathcal{S}_\mathcal{U})\left ( \frac{p}{1-q} \right)
    \end{cases}.
\end{equation*}

Clearly the first two cases are non-negative, so we consider the third case.  

Because $\mathcal{S}_\mathcal{U} > q$, we have $\frac{\mathcal{S}_\mathcal{U}}{q}>1>\frac{1-\mathcal{S}_\mathcal{U}}{1-q}$ and 
\begin{align*}
    f'\left ( \frac{\mathcal{S}_\mathcal{U}}{q}  \right ) &>f'\left ( \frac{1-\mathcal{S}_\mathcal{U}}{1-q}  \right )\\
    f'\left ( \frac{\mathcal{S}_\mathcal{U}}{q}  \right )-f'\left ( \frac{1-\mathcal{S}_\mathcal{U}}{1-q}  \right )&>0.
\end{align*}

We can no prove the fourth case is positive.

\begin{align*}
    \mathcal{S}_\mathcal{U}&>(1-\mathcal{S}_\mathcal{W})\left ( \frac{q}{1-p} \right)\\
    \mathcal{S}_\mathcal{U}-p\mathcal{S}_\mathcal{W}&>q-q\mathcal{S}_\mathcal{W}\\
    \mathcal{S}_\mathcal{U}-p\mathcal{S}_\mathcal{W}-q\mathcal{S}_\mathcal{U}&>q-p\mathcal{S}_\mathcal{W}-q\mathcal{S}_\mathcal{U}\\
    \frac{\mathcal{S}_\mathcal{U}}{q}&>\frac{1-\mathcal{S}_\mathcal{W}-\mathcal{S}_\mathcal{U}}{1-p-q}\\
    f'\left (\frac{\mathcal{S}_\mathcal{U}}{q}\right )&>f'\left ( \frac{1-\mathcal{S}_\mathcal{W}-\mathcal{S}_\mathcal{U}}{1-p-q}\right )\\
    f'\left (\frac{\mathcal{S}_\mathcal{U}}{q}\right ) - f'\left ( \frac{1-\mathcal{S}_\mathcal{W}-\mathcal{S}_\mathcal{U}}{1-p-q}\right )&>0
\end{align*}

This shows that \eqref{undesirable_derivative} is non-negative and \eqref{closed_form} is increasing in $\tilde{y}_i$ for $i\in \mathcal{U}$.

\newpage 
\subsection{Proofs of Theorem \ref{theorem_PAC} (PAC generalization bounds for Verifier)}\label{APP_PAC}

Let us define $\mathcal{D}_{i}$ as the distribution of the data $\mathbf{x}$ conditioned on the event that it is drawn from class $i$. 
We define a loss function $\ell:\{0,1\}\times[0,1]\rightarrow\mathbb{R}$ as follows:
\begin{equation}
    \ell (z,v) = zl(v)+(1-z)l(1-v),
\end{equation}
where $l$ is some differentiable function (e.g., $\log()$ which would lead to the cross-entropy loss).  Furthermore, we assume that the output of the loss $\ell$ is upper bounded by a constant $B_\ell$ and is Lipschitz. The verifier output of $V(\mathbf{x},\breve{\mathbf{x}})$ estimates probability that $\mathbf{x}$ and $\breve{\mathbf{x}}$ belong to the same class. 

Using this loss, we now define the true risk $R(V)$ of a verifier $V$ as 
\begin{equation}
    R(V) = \underbrace{\frac{1}{k(k -1)} \sum_{i \neq j}\mathop{\mathbb{E}}_{\mathbf{x}^{(i)}\sim\mathcal{D}_i, \mathbf{{x}}^{(j)}\sim\mathcal{D}_j}[\ell(0,V(\mathbf{x}^{(i)},{\mathbf{x}}^{(j)}))]}_{R^{(\text{diff})}(V)}+\underbrace{\frac{1}{k} 
 \sum_{i=1}^{k}
\mathop{\mathbb{E}}_{(\mathbf{x},\breve{\mathbf{x}})\sim\mathcal{D}_i}[\ell(1,V(\mathbf{x},\breve{\mathbf{x}}))]}_{R^{(\text{same})}(V)}.
\end{equation}
The verifier faces two types of inputs that it should be able to distinguish: (a) pairs of inputs that can come from the same class (i.e., $\mathbf{x},\breve{\mathbf{x}} \sim \mathcal{D}_i$ for some class $i$) and (b) pairs of inputs that can belong to different classes (i.e., $\mathbf{x}^{(i)} \sim \mathcal{D}_i$ and ${\mathbf{x}^{(j)}} \sim \mathcal{D}_j$ for some pair of classes $i\neq j$). This formulation of risk assigns equal value to identifying pairs from the same class and pairs form different classes because both of  $R^{(\text{diff})}(V)$ (accuracy on pairs form different classes) and $R^{(\text{same})}(V)$ (accuracy on pairs form the same class) are normalized by dividing by the total number of terms in the sum. Specifically, we normalize the total risk for misclassifying pairs from different classes by $k(k-1)$, which is the number of distinct ordered pairs of classes we can form out of $k$ classes. Similarly, we normalize the total risk of misclassifying pairs from same classes by $k$.  Furthermore, both $R^{(\text{diff})}(V)$  and $R^{(\text{same})}(V)$ assign equal importance to each possible type of combination (which class the first element of the pair comes form and which class the second element of the pair comes from).

To calculate our empirical risk we will assume we are given $k$  sets $\mathcal{S}^{(i)}$ $1\leq i\leq k$, each containing $n/k$ samples drawn independently from the corresponding $\mathcal{D}_i$ as defined above.  We index these sets as follows:
\begin{align}
    \mathcal{S}^{(i)} = \{ \mathbf{x}^{(i)}_{(q)}\}^{n/k}_{q = 1}, \quad i=1,2,\ldots, k.
\end{align}
We define the entire dataset $\mathcal{S}$ as
\begin{align}
\mathcal{S} =  \bigcup\limits_{i = 1}^{k}\mathcal{S}^{(i)}
\end{align} 
We define our empirical risk for training the verifier over the set $\mathcal{S}$ as follows:
\begin{equation}
    \hat{R}_{\mathcal{S}}(V) = \underbrace{\frac{1}{k(k -1)} \sum_{i \neq j} \frac{1}{\left(\frac{n}{k}\right)^2} \sum_{q = 1}^{\frac{n}{k}} \sum_{r = 1}^{\frac{n}{k}}\ell(0,V(\mathbf{x}^{(i)}_q,{\mathbf{x}}^{(j)}_r))}_{\hat{R}_{\mathcal{S}}^{(\text{diff})}(V)}+\underbrace{\frac{1}{k} 
 \sum_{i=1}^{k} \frac{1}{\left(\frac{n}{k}\right)^2 - n}
\sum_{q = 1}^{\frac{n}{k}}\sum^{\frac{n}{k}}_{r = 1, r \neq q}\ell(1,V(\mathbf{x}^{(i)}_q,{\mathbf{x}}^{(i)}_r))}_{\hat{R}^{(\text{same})}_{\mathcal{S}}(V)}.
\end{equation}
where $\hat{R}_{\mathcal{S}}^{(\text{diff})}(V)$ denotes the empirical risk of the verifier on inputs from different classes; and $\hat{R}_{\mathcal{S}}^{(\text{same})}(V)$ denotes the empirical risk of the verifier on inputs from the same class. It is straightforward to verify that $\hat{R}_{\mathcal{S}}(V)$ is an unbiased estimator of the true risk $R(V)$, i.e., $\mathbb{E}(\hat{R}_{\mathcal{S}}(V))= {R}(V)$.

Let us define worst case generalization gap for a given dataset $\mathcal{S}$ as 
\begin{equation}
    \phi(\mathcal{{S}}) = \sup_{V \in \mathcal{V}} \left | \mathrm{R}(V) - \hat{\mathrm{R}}_{\mathcal{S}}(V) \right |,
    \label{eq: gen-gap-Rademacher}
\end{equation}
where $\mathcal{V}$ denotes the hypothesis class from which the verifier $V(\cdot)$ is selected. 
To bound this generalization gap, we will use the notion of Rademacher complexity which measures the correlation between the function class and the random labels to upper bound the generalization gap \cite{mohri2018foundations}. The Rademacher complexity of a hypothesis class over a particular data set is formally defined as: 
 \begin{definition}
     The empirical Rademacher complexity of a function class $\mathcal{F}$ with respect to the set $\mathcal{\check{S}} = \{\mathbf{a}_i\}_{i=1}^n$ is given by the following equation:
     \begin{equation}
         \mathfrak{R}_{\mathcal{\check{S}}}(\mathcal{F})=\frac{1}{n}\underset{\sigma}{\mathbb{E}}\left[\sup_{f\in\mathcal{F}} \sum_{i=1}^n \sigma_i f(\mathbf{a}_i)\right],
    \end{equation}
     where $\sigma_{i}$'s are i.i.d. Rademacher random variables, i.e., $\Pr(\sigma_i = 1) = \Pr(\sigma_i = -1) = \frac{1}{2}$.
 \end{definition}

In the following steps, we upper bound the generalization gap in \eqref{eq: gen-gap-Rademacher} as using Rademacher complexity. 
We first bound the generalization gap using triangle inequality as follows:
\begin{align}
    \phi(\mathcal{S}) =& \sup_{V\in\mathcal{V}}\left | R^{(\text{diff})}(V)+R^{(\text{same})}(V) - \hat{R}^{(\text{diff})}_\mathcal{S}(V) -\hat{R}^{(\text{same})}_\mathcal{S}(V) \right |\\
    \leq & \sup_{V\in\mathcal{V}}\left | R^{(\text{diff})}(V) - \hat{R}^{(\text{diff})}_\mathcal{S}(V) \right | +\sup_{V\in\mathcal{V}}\left | R^{(\text{same})}(V) - \hat{R}^{(\text{same})}_\mathcal{S}(V) \right |
\end{align}
The above bound first decomposes the generalization gap into the sum of two generalization gaps, the first over the pair of samples coming from different classes; and the second over the samples drawn from the same class. 
To proceed we will need a few additional definitions: we define $\mathcal{D}_i \times  \mathcal{D}_j$ to represent the distribution over pairs $(\mathbf{x}^{(i)},\mathbf{x}^{(j)})$ where $\mathbf{x}^{(i)}$ is drawn from $\mathcal{D}_i$ and $\mathbf{x}^{(j)}$ is drawn from $\mathcal{D}_j$ independently.
We also define the sets
\begin{equation}
    \mathcal{S}^{(i)} \times \mathcal{S}^{(j)} = \begin{cases} 
          \{(\mathbf{x}^{(i)}_{(q)},{\mathbf{x}}^{(j)}_{(r)}) \}_{1\leq q,r\leq k} & i\neq j \\
           \{(\mathbf{x}^{(i)}_{(q)},{\mathbf{x}}^{(j)}_{(r)}) \}_{1\leq q,r\leq k,q\neq r} & i = j
       \end{cases},
\end{equation} and enumerate the elements of each set by $\mathcal{S}^{(i)} \times \mathcal{S}^{(j)} = \{u^{i,j}_q\}_{q=1}^{(n/k)^2}$ when $i\neq j$. When $i=j$ the enumeration takes the form $\mathcal{S}^{(i)} \times \mathcal{S}^{(i)} = \{u^{i,i}_q\}_{q=1}^{(n/k)^2-n}$.

Using our definitions of true and empirical risk, we can now upper bound the above sum as follows, 
\begin{align}
\phi(\mathcal{S}) \leq& \sup_{V \in \mathcal{V}} \frac{1}{k(k-1)} \Biggl|\sum_{i \neq j} \left[\mathbb{E}_{\mathbf{x}^{(i)} \in \mathcal{D}_i,\mathbf{x}^{(j)} \in \mathcal{D}_j}\left[\ell(V(\mathbf{x}^{(i)},\mathbf{x}^{(j)}),0)\right] - \left(\frac{k}{n} \right)^2 \sum_{q = 1}^{\frac{n}{k}} \sum_{r = 1}^{\frac{n}{k}} \ell(V(\mathbf{x}_{(q)}^{(i)},\mathbf{x}_{(r)}^{(j)}),0)\right] \Biggr| \nonumber \\
& + \sup_{V \in \mathcal{V}} \frac{1}{k} \Biggl|\sum_{k = 1}^{j} \left[\mathbb{E}_{\mathbf{x},\breve{\mathbf{x}} \in \mathcal{D}_i}\left[\ell(V(\mathbf{x},\breve{\mathbf{x}}),1)\right] -  \sum_{q = 1}^{\frac{n}{k}} \sum_{r = 1, r \neq q}^{\frac{n}{k}} \ell(V(\mathbf{x}_{(q)}^{(i)},\mathbf{x}_{(r)}^{(i)}),1)\right] \Biggr|
\nonumber \\
{\leq}& \frac{1}{k(k-1)}\sum_{i \neq j} \sup_{V \in \mathcal{V}} \Biggl| \mathbb{E}_{u \sim \mathcal{D}_i \times  \mathcal{D}_j} \ell(V(u),1) - \left(\frac{k}{n}\right)^2 \sum_{q = 1}^{\left(\frac{n}{k}\right)^2} \ell(V(u^{(i,j)}_q), 1) \Biggr| \nonumber \\
& + \frac{1}{k}\sum_{i=1}^k \sup_{V \in \mathcal{V}} \Biggl| \mathbb{E}_{u \sim \mathcal{D}_i \times  \mathcal{D}_i} \left[\ell(V(u),1)\right] - \left(\frac{k}{n}\right)^2 \sum_{q = 1}^{\left(\frac{n}{k}\right)^2-n} \ell(V(u^{(i,i)}_q), 1) \Biggr| \label{supremum_division}
\end{align}
where the second inequality follows by bounding the absolute value of a sum by the sum of the absolute values (across both the ``diff" and ``same" terms).  

We now use the standard Rademacher complexity PAC-bound \cite{mohri2018foundations, bartlett2002rademacher} on each of the supremums in (\ref{supremum_division}).  The result is that for any $\delta\in(0,1)$,formulation the following holds with probability $1-\delta$ over the choice of $\mathcal{S}$:
\begin{align}
 \phi(\mathcal{S}){\leq}& \frac{1}{k(k-1)}\sum_{i \neq j} \left[2 \mathfrak{R}_{\mathcal{S}^{(i)} \times \mathcal{S}^{(j)}} (\mathcal{V}) + \frac{6k B_{\ell}}{n} \sqrt{\frac{\log(2/\delta)}{2}} \right] + \frac{1}{k} \sum_{i = 1}^{k}  \left[2 \mathfrak{R}_{\mathcal{S}^{(i)} \times \mathcal{S}^{(i)}} (\mathcal{V}) + \frac{6k B_{\ell}}{\sqrt{n^2 -k^2n}} \sqrt{\frac{\log(2/\delta)}{2}} \right]
\nonumber \\
 =& \frac{2}{k(k-1)}\sum_{i \neq j} \mathfrak{R}_{\mathcal{S}^{(i)} \times \mathcal{S}^{(j)}}(\mathcal{V}) + \frac{2}{k} \sum_{i=1}^k\mathfrak{R}_{\mathcal{S}^{(i)} \times \mathcal{S}^{(i)}}(\mathcal{V}) + \frac{6k B_{\ell}}{n} \sqrt{\frac{\log(2/\delta)}{2}} + \frac{6k B_{\ell}}{\sqrt{n^2 -k^2n}} \sqrt{\frac{\log(2/\delta)}{2}}
\nonumber \\
 \leq& \frac{2}{k(k-1)}\sum_{i \neq j} \mathfrak{R}_{\mathcal{S}^{(i)} \times \mathcal{S}^{(j)}}(\mathcal{V}) + \frac{2}{k} \sum_{i=1}^k\mathfrak{R}_{\mathcal{S}^{(i)} \times \mathcal{S}^{(i)}}(\mathcal{V}) + \frac{12k B_{\ell}}{\sqrt{n^2 -k^2n}} \sqrt{\frac{\log(2/\delta)}{2}}\label{expression:proof}
\end{align}
where the final inequality comes form replacing $\frac{6k B_{\ell}}{n} $ with the larger  $\frac{6k B_{\ell}}{\sqrt{n^2 -k^2n}}$.
Equation \eqref{expression:proof} can be interpreted as the sum of three terms: the first term is the average Rademacher complexity over the datasets corresponding to pairs which are drawn from different classes; the second term is the average Rademacher complexity over the datasets corresponding to pairs which are drawn from same classes; the third term is a standard term which shows the dependence on $\delta$ (as well as $n,k$). 

We now apply the bound on empirical Rademacher complexity
\begin{equation}
    \mathfrak{R}_\mathcal{Q}(V)=\mathcal{O}\left ( |\mathcal{Q}|^{-1/d'} \right )
\end{equation}
with $d'$ the dimension of the elements of $\mathcal{Q}$ \cite{gottlieb2016adaptive}. To apply this we will recall the dimension of the elements of $\mathcal{S}^{(i)} \times \mathcal{S}^{(j)}$ is $2d$, and $|\mathcal{S}^{(i)} \times \mathcal{S}^{(j)}| = \left ( \frac{n}{k} \right )^2$ when $i\neq j$, and $|\mathcal{S}^{(i)} \times \mathcal{S}^{(i)}| = \left ( \frac{n}{k} \right )^2-n = \frac{n^2-k^2n}{k^2}$.  Applying our Rademacher complexity bound yields
\begin{align}
    \phi(\mathcal{S}) \leq& \frac{2}{k(k-1)}\sum_{i \neq j} \mathcal{O}\left ( \left (\frac{k}{n}\right )^{1/d} \right ) + \frac{2}{k} \sum_{i=1}^k\mathcal{O}\left ( \left (\frac{k}{\sqrt{n^2-k^2n}} \right )^{1/d} \right ) + \frac{12k B_{\ell}}{\sqrt{n^2 -k^2n}} \sqrt{\frac{\log(2/\delta)}{2}}\\
    =&2 \mathcal{O}\left ( \left (\frac{k}{n}\right )^{1/d} \right ) + 2\mathcal{O}\left ( \left (\frac{k}{\sqrt{n^2-k^2n}} \right )^{1/d} \right ) + \frac{12k B_{\ell}}{\sqrt{n^2 -k^2n}} \sqrt{\frac{\log(2/\delta)}{2}}\\
    \leq& 4 \mathcal{O}\left ( \left (\frac{k}{\sqrt{n^2-k^2n}} \right )^{1/d} \right ) + \frac{12k B_{\ell}}{\sqrt{n^2 -k^2n}} \sqrt{\frac{\log(2/\delta)}{2}}.
    \label{eq: final-PAC-bound}
\end{align}

The bound in \eqref{eq: final-PAC-bound} is our final PAC bound true with probability $1-\delta$.  However, we expect the $\delta$ containing term to be dominated by the other term because $(\frac{k}{\sqrt{n^2-k^2n}})<1$ and $d$ is expected to be much larger than $1$.

\subsection{Additional Implementation Details}\label{APP_implementation}

In this section we give additional details on how we implemented our methods to create the experimental results found in this paper.

\subsubsection{Data Set and Cost Function Details}\label{APP_datasets}

Here we give additional description of each data set and the corresponding the cost functions $d_\mathcal{X}$ used in our experiments.  As noted in Section \ref{sec:creation} we must ensure $d_\mathcal{X}$ is differentiable.  When dealing with categorical features costs are by nature discrete (and not differentiable).  We show how we were able to write these costs in a differentiable form.  Suppose $\mathbf{v}\in\mathbb{R}^\ell$ is a one-hot encoding of a categorical feature and define the \textit{transition cost matrix} $A$ such that $A_{i,j}$ as the cost of changing from category $i$ to category $j$.  Then $\mathbf{z}^T A \tilde{\mathbf{z}}$ represents the costs of changing this categorical feature and is differentiable in $\mathbf{z}$.

\textbf{Adult Income Prediction Dataset}:  \citep{adult_DS}  This widely used data set contains information from the 1994 U.S. census, with individuals labelled by whether their annual income was over \$50,000 ($\sim$\$100,000 in 2023 adjusted for inflation).  We define our target set $T$ as over 80\% probability high income. Our actionable set allows changes in job type, education and number of hours worked with all other attributes immutable.  The cost function $d_\mathcal{Y}$ includes the expected number of years to improve education (e.g. two years to go from associate's degree to bachelors degree), a one-year cost to change employer type and the 2-norm of the change in hours worked per week (weighted so 3 hours per week is equivalent to a year spent on education).  Here Trustworthy Actionable Perturbations suggest the best way to improve an individuals odds of making a large income with the least time and effort.

Specifically $d_\mathcal{X}$ is the sum cost from changes (1) hours worked per week (2) change in employment type (3) change in education and (4) change in field of work.

The cost from a change in hours is given by $\frac{\Delta h^2}{10}$ where $\Delta h$ is the change in weekly hours worked.  This will mean $3$ extra hours of work are approximately equivalent to one year of schooling.

The cost from a change in employer (the options are government, private, self-employed and other) is always $1$ (equivalent to a year spent on education).

The possible levels of education are (1) any schooling, (2) High School Degree, (3) Professional Degree, (4) some college, (5) Associate's Degree, (6) Bachelors Degree, (7) Master's Degree, (8) Doctorate Degree.  The cost transition matrix associated with the level of education (as ordered above) is 
\begin{equation}
    A_{\text{Education}} = \begin{bmatrix}
0&2&10&3&4&6&8&11\\
L&0&8&1&2&4&6&9\\
L&L&0&L&L&L&2&5\\
L&L&7&0&1&3&5&8\\
L&L&6&L&0&2&4&7\\
L&L&4&L&L&0&2&5\\
L&L&4&L&L&L&0&3\\
L&L&4&L&L&L&L&0
\end{bmatrix},
\end{equation}
where $L$ is a large number meant to prevent suggestions that lead to a decrease in education, which is impossible (we use $L = 1,000$).  These numbers represent the expected number of years required to gain the specified degree (i.e. the cost of going from a high school degree to a bachelors degree is $A_{2,6}=4$).

Finally the options for fields of work are (1) Service, (2) Sales, (3) Blue-Collar (4) White Collar, (5) Professional, (6) Other.  The cost transition matrix associated with the level of education (as ordered above) is 
\begin{equation}
    A_{\text{Profession}} = \begin{bmatrix}
0&1&2&3&4&1\\
1&0&1&2&3&1\\
1&1&0&1&2&1\\
1&1&1&0&1&1\\
1&1&1&1&0&1\\
1&1&1&1&1&0
\end{bmatrix}.
\end{equation}
This represents a cost of $1$ for any change

\textbf{Law School Success Prediction Dataset}:  \citep{law_DS}  This data set contains demographic information and  academic records for over 20,000 law school students labelled by whether or not a student passed the BAR exam.  Our target set is an 85\% chance of passing the BAR. To create $\mathcal{A}(\mathbf{x)}$, we suppose the law school performance is merely a projection that can be changed through more studying, allowing us to change the law school grades and the location where the students take the BAR. The cost function $d_\mathcal{X}$ sums the increase in grades and the physical distance travelled to take the BAR where moving to an adjacent region (e.g. Far West to North West) is weighted the same as increasing grades by one standard deviation.

Specifically $d_\mathcal{Y}$ sums the increase in grades and the physical distance travelled to take the BAR where moving to an adjacent region (e.g. Far West to North West) is weighted the same as increasing grades one standard deviation.  This set up returns the optimal combination of studying harder and moving location to take the BAR.In this data set $d_\mathcal{X}$ is sum of the change in grades (in standard deviations from the mean) and distance traveled.  The country was divided into eight regions: (1) Far West, (2) Great Lakes, (3) Mid-South, (4) Mountain West, (5) Mid-West, (6) North East, (7) New England, (8) North West.  We use the transition cost matrix
\begin{equation}
    A_{\text{Region}} = \begin{bmatrix}
0&3&4&1&2&6&5&1\\
3&0&1&2&1&2&1&3\\
4&1&0&2&1&2&1&5\\
1&2&2&0&1&4&3&2\\
2&1&1&1&0&3&2&3\\
6&2&2&4&3&0&1&5\\
5&1&1&3&2&1&0&5\\
1&3&5&2&3&5&5&0
\end{bmatrix}
\end{equation}
Moves to adjacent regions result in a cost of $1$, while the highest cost of $6$ is incurred by moving from Far West  to New England or back.

\textbf{Diabetes Prediction Dataset}: \citep{diabetes_DS}  This data set contains information on the demographics, health conditions and health habits of 250,000 individuals labelled by whether an individual is diabetic extracted from the Behavioral Risk Factor Surveillance System (BRFSS), a health-related telephone survey that is collected annually by the CDC.. We define $\mathcal{A}(\mathbf{x})$ to allow changes in health habits, BMI, education and income.  We use a weighted 2-norm for $d_\mathcal{X}$ to represent the relative difficulty of making changes.  For example, starting to get regular physical activity is weighted the same as dropping one BMI point.  Increasing education, income and health insurance were weighted as more difficult that simply adjusting health habits.

\textbf{German Credit Dataset}: \citep{germancredit_DS}  This commonly used data set contains information on 1,000 loan applications in Germany labelled by their credit risk.  The actionable set allows for changes in the loan request (time and size) as well as the funds in the applicants checking and savings account and whither the applicant has a telephone.  The target set $T$ is a greater than 80\% of being a good credit risk. The cost function $d_\mathcal{Y}$ is the direct measuring the total difference in Deutsche Marks (DM) between all elements of the application.  No cost was assigned to closing empty accounts.  The change in length of loan is converted to DM through the individual's monthly disposable income.  Finally we set a flat cost of 50DM to acquire a telephone. 

\subsubsection{Model Details}\label{APP:models}

We used fully connected feed forward  neural networks. Each network used 3 hidden layers with ReLu activation functions between each layer.  We tuned the parameters of the neural networks until we achieved accuracy on par with common tree based classifiers (random forests and histogram boosted trees).  Accuracy results are presented in table \ref{tab:trees_vs_nn}.  For all data sets except the German Credit data set each hidden layer had $60$ nodes.  The German Credit data set required $120$ nodes per layer.  Additionally, for the German Credit data set only, we used dropout regularization of $20\%$ on each hidden layer.  We trained these models using the ADAM optimizer to minimize cross entropy loss.  We used an $80-10-10$ train-validate-test data split and implemented early stopping with the validation data. All Trustworthy Actionable Perturbations, counterfactuals and adversarial examples were created for the testing data.  We used identical architecture for $V$ as $M$, except for doubling the input size. Accuracy data may be found in table \ref{tab:data_summary}.

\begin{center}
\begin{tabular}{ |c|c|c|c|c| } 
 \hline
  & Adult Income & Law School Success & Diabetes Prediction & German Credit\\ 
 \hline\hline
 Random Forest & 73\% & 64\% & 62\% & 74\% \\ 
 \hline
 Histogram Gradient Boosted Trees & 81\% & 77\% & 75\% & 69\% \\ 
 \hline
 Neural Network & 80\% & 77\% & 75\% & 75\% \\
 \hline
\end{tabular}\label{tab:trees_vs_nn}
\end{center}

We also tested the calibration of our networks by calculating the \textit{expected calibration error} (ECE) \cite{ECE}.  We used 15 bins and record the results in table \ref{tab:calibration}

\begin{center}
\begin{tabular}{ |c|c|c|c|c| } 
 \hline
  & Adult Income & Law School Success & Diabetes Prediction & German Credit\\ 
 \hline
 ECE (15 bins) & 16\% & 15\% & 7\% & 21\% \\ 
 \hline
\end{tabular}\label{tab:calibration}
\end{center}

\subsubsection{Objective Function Details}\label{APP:objective}

In our implementation we formulated the actionablility penalty term $b$ as 
\begin{equation}
    b(\tilde{\mathbf{x}}) = G\left (\sum_{i=1}^m \max\{0,\tilde{\mathbf{x}}_i-u_i\} + \max\{0,l_i-\tilde{\mathbf{x}}_i\}\right )
\end{equation}
with $G$ a sufficiently large constant.
We formulated our coherence penalty term $p$ as 
\begin{equation}
    p(\tilde{\mathbf{x}}) = P\sum_{i=1}^{C}\left ( 1 - \sum_{j\in \mathcal{C}_i}  \tilde{\mathbf{x}}_j  \right )^2,
\end{equation}
with $P$ another appropriately large constant. The conditioner function $cond$ simply rounded integer and Boolean values to the nearest integer value.  For one-hot encoded features categorical features, the category with the largest value set to one and all other categories set to zero.

\subsubsection{Additional Experimental Results}\label{APP:results}

\begin{figure}[t]
\begin{center}
\includegraphics[width = 1\textwidth]{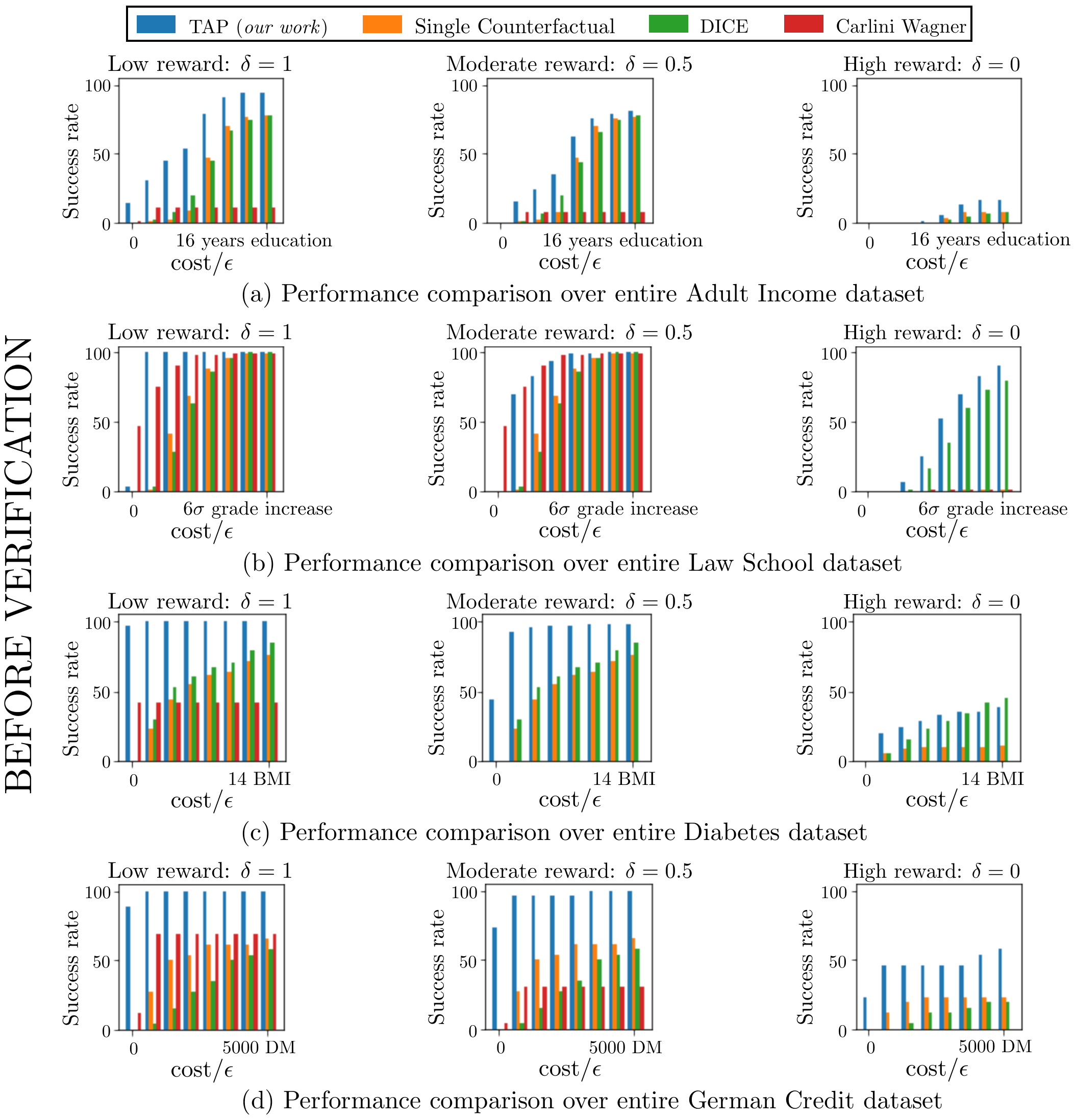}
\caption{\textbf{Performance comparison over entire datasets before verification:} The graphs show average success rate for moving individuals within a variety of distances ($\delta$) to the target set.  The y-axis shows the percentage of individuals  within the goal distance, and the x-axis, represents different costs ($\epsilon$ values) to achieve the goal.  These values were obtained before applying the verification procedure.}
\label{fig:full_unverified_comparison}
\end{center}
\end{figure}

Here we show success bar charts similar to those found in figure \ref{fig:full_unverified_comparison} compare the efficacy of Trustworthy Actionable Perturbations, counterfactuals \cite{original_CF,Diverse_CF} and adversarial examples from the Carlini Wagner $\ell_2$ attack \cite{carliniwagner_AA} for all data sets.  These are similar to Figure \ref{fig:full_comparison}, but include all data sets and an increased number of cost ($\epsilon$) values.

Each bar chart refers to a particular data set and desired distance $\delta$ to the target set $T$.  Inside of each chart, the bars show the percentage of individuals that a method was able to successfully move inside the goal $\delta$ at a variety of costs $\epsilon$.  Figure \ref{fig:full_unverified_comparison} shows data before the verification procedure has been performed and \ref{fig:full_unverified_comparison} shows the data after all .  In these tests, the Trustworthy Actionable Perturbations (in {\color{blue}blue}) outperform the counterfactuals  (in {\color{green}green} and {\color{orange}orange}) in nearly all cases except for when both methods achieved $100\%$ success or the very high-cost (large $\epsilon$) high reward ($\delta=0$) scenarios.  Carlini Wagner attacks ({\color{red}red}) are only effective at larger $\delta$ values because they are designed to move a data point just barely inside the target class.  The Carlini Wagner attacks are not required to be actionable (or even feasible), so they do not constitute useful advise.  The verifier is able to recognize that these adversarial examples are untrustworthy in all cases.

\begin{figure}[t]
\begin{center}
\includegraphics[width = 1\textwidth]{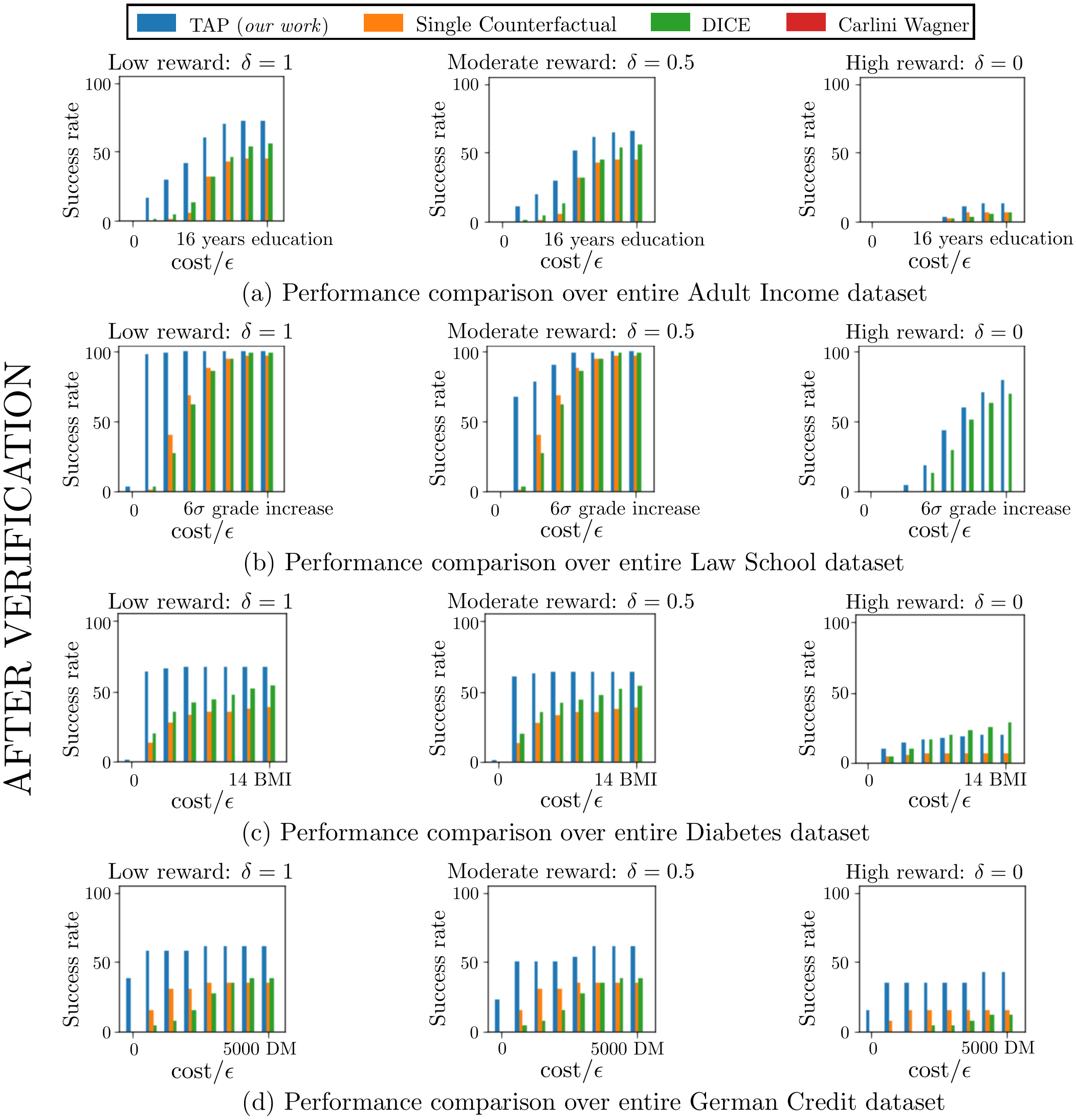}
\caption{\textbf{Performance comparison over entire datasets after verification:} The graphs show average success rate for moving individuals within a variety of distances ($\delta$) to the target set.  The y-axis shows the percentage of individuals  within the goal distance, and the x-axis, represents different costs ($\epsilon$ values) to achieve the goal.  These values were obtained after applying the verification procedure with a $10\%$ chance of eliminating valid inputs.}
\label{fig:full_verified_comparison}
\end{center}
\end{figure}

\end{document}